\documentclass{ieeeaccess}
\usepackage{cite}
\usepackage{amsmath,amssymb,amsfonts}
\usepackage{algorithmic}
\usepackage{graphicx}
\usepackage{textcomp}
\usepackage{multirow}
\usepackage{hyperref}
\usepackage{soul}

\usepackage{bm}
\makeatletter
\AtBeginDocument{\DeclareMathVersion{bold}
\SetSymbolFont{operators}{bold}{T1}{times}{b}{n}
\SetSymbolFont{NewLetters}{bold}{T1}{times}{b}{it}
\SetMathAlphabet{\mathrm}{bold}{T1}{times}{b}{n}
\SetMathAlphabet{\mathit}{bold}{T1}{times}{b}{it}
\SetMathAlphabet{\mathbf}{bold}{T1}{times}{b}{n}
\SetMathAlphabet{\mathtt}{bold}{OT1}{pcr}{b}{n}
\SetSymbolFont{symbols}{bold}{OMS}{cmsy}{b}{n}
\renewcommand\boldmath{\@nomath\boldmath\mathversion{bold}}}
\makeatother

\def\BibTeX{{\rm B\kern-.05em{\sc i\kern-.025em b}\kern-.08em
    T\kern-.1667em\lower.7ex\hbox{E}\kern-.125emX}}

\begin{document}
\history{Date of publication xxxx 00, 0000, date of current version xxxx 00, 0000.}
\doi{10.1109/ACCESS.2023.1120000}

\title{Fighting Against the Repetitive Training and Sample Dependency Problem in Few-shot Named Entity Recognition}
\author{\uppercase{CHANG TIAN}\authorrefmark{1},
\uppercase{WENPENG YIN}\authorrefmark{2}, 
\uppercase{Dan Li}\authorrefmark{3}, 
and MARIE-FRANCINE MOENS\authorrefmark{1}}

\address[1]{LIIR, Department of Computer Science, KU Leuven, 3000 Leuven, Belgium}
\address[2]{AI4Research Lab, Pennsylvania State University, State College, PA 16801, United States}
\address[3]{Elsevier, Amsterdam 1043 NX, Netherlands}
\tfootnote{This work was supported in part by the China Scholarship Council, and Marie-Francine Moens is supported by the ERC Advanced Grant H2020-ERC-2017-ADG 788506.}

\markboth
{Author \headeretal: Preparation of Papers for IEEE TRANSACTIONS and JOURNALS}
{Author \headeretal: Preparation of Papers for IEEE TRANSACTIONS and JOURNALS}

\corresp{Corresponding author: Chang Tian (e-mail: chang.tian@kuleuven.be).}

\begin{abstract}
Few-shot named entity recognition (NER) systems recognize entities using a few labeled training examples. The general pipeline consists of a span detector to identify entity spans in text and an entity-type classifier to assign types to entities. Current span detectors rely on extensive manual labeling to guide training. Almost every span detector requires initial training on basic span features followed by adaptation to task-specific features. This process leads to repetitive training of the basic span features among span detectors. Additionally, metric-based entity-type classifiers, such as prototypical networks, typically employ a specific metric that gauges the distance between the query sample and entity-type referents, ultimately assigning the most probable entity type to the query sample. However, these classifiers encounter the sample dependency problem, primarily stemming from the limited samples available for each entity-type referent. To address these challenges, we proposed an improved few-shot NER pipeline. First, we introduce a steppingstone span detector that is pre-trained on open-domain Wikipedia data. It can be used to initialize the pipeline span detector to reduce the repetitive training of basic features. Second, we leverage a large language model (LLM) to set reliable entity-type referents, eliminating reliance on few-shot samples of each type. Our model exhibits superior performance with fewer training steps and human-labeled data compared with baselines, as demonstrated through extensive experiments on various datasets. Particularly in fine-grained few-shot NER settings, our model outperforms strong baselines, including ChatGPT. We will publicly release the code, datasets, LLM outputs, and model checkpoints.
\end{abstract}


\titlepgskip=-21pt

\maketitle

\section{Introduction}
\label{sec:introduction}
\begin{figure}[th!]
\includegraphics[width=0.45\textwidth]{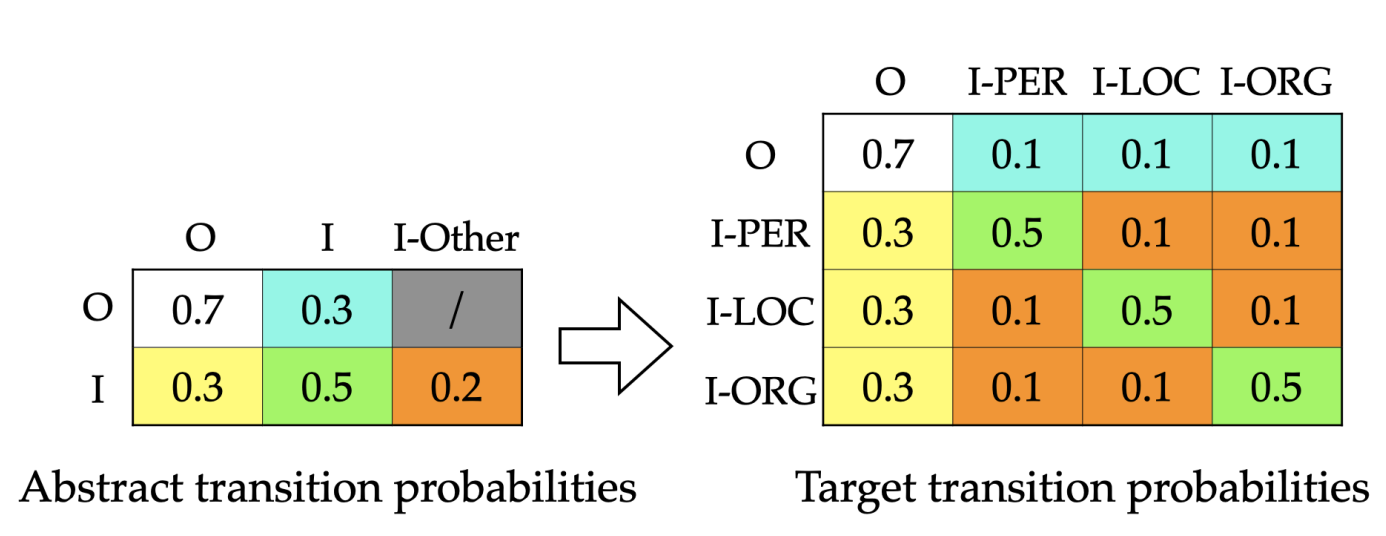}
\centering
\caption{An example of token-wise label dependency. The label dependency is shown with transition probabilities between entity labels; for example, the probability from I to O is 0.3 in the left table. This figure is from the work~\cite{08}, which assumes that an abstract transition probability is
evenly split into related target transitions. However, this does not match reality.}
\label{img:label_dependency}
\end{figure}

\begin{figure}[th!]
\includegraphics[width=0.45\textwidth]{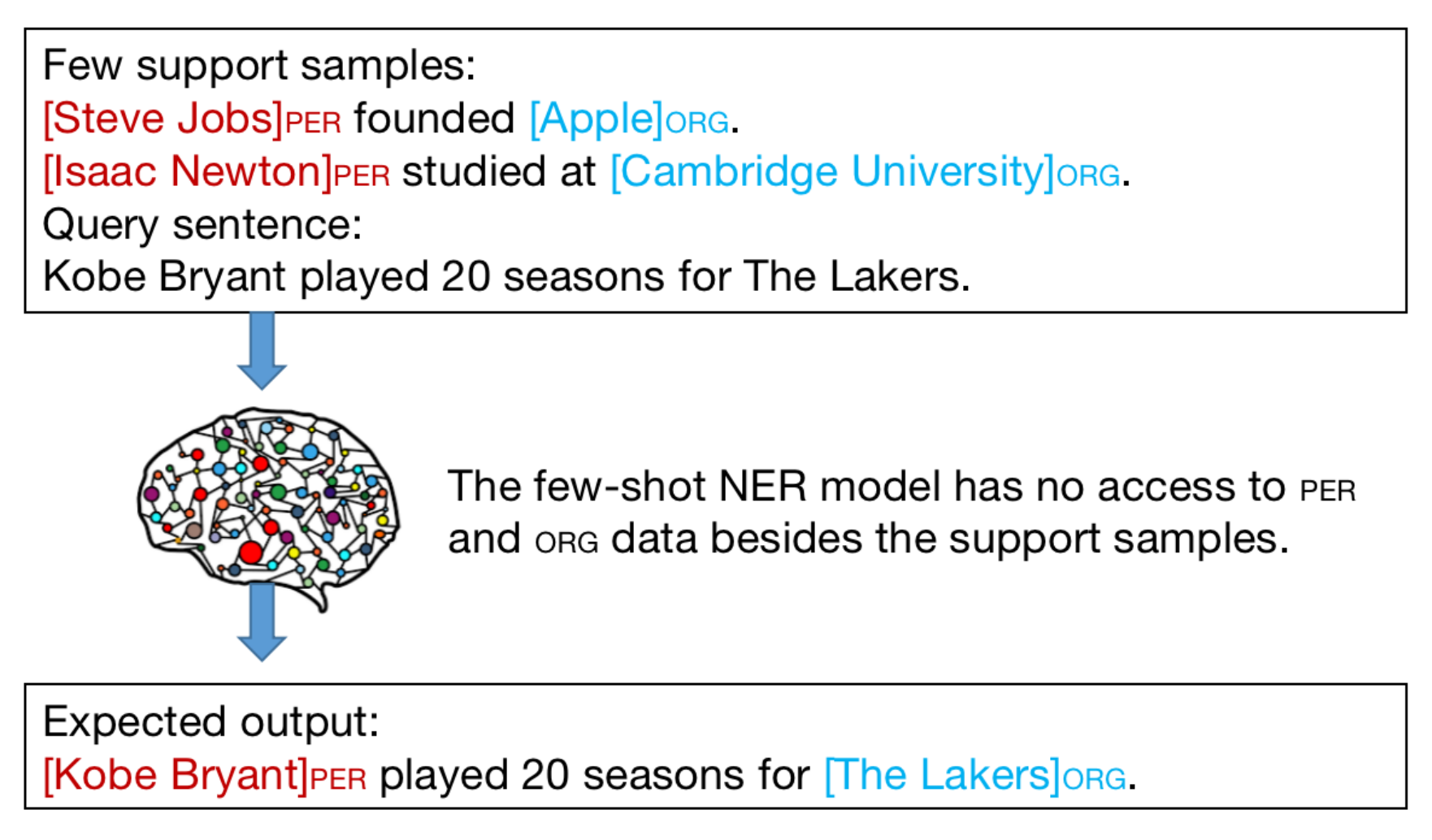}
\centering
\caption{2-way 2-shot NER task. Model detects new entities using two shots per target type. Colors represent different entity types.}
\label{img:22sample}
\end{figure}

Named Entity Recognition (NER)~\cite{01},~\cite{02},~\cite{03} involves detecting and classifying text span into predefined types. Traditional NER approaches struggle when faced with unlabeled texts and only a few labeled examples (\textit{support examples}) for each target entity type. In real-world applications, annotated data for training NER models can be scarce~\cite{04}, particularly in specific domains or languages. 
Few-shot NER models~\cite{04,05} are designed to recognize target entities with minimal labeled examples, demonstrating effective performance even with scarce training data.
Figure~\ref{img:22sample} illustrates a few-shot NER task. 
Two types of few-shot NER systems~\cite{06} exist: pipeline and end-to-end systems. End-to-end NER systems~\cite{07},~\cite{08},~\cite{09},~\cite{10} label each token in a sentence with both a span tag (e.g., "I", "O") and an entity type tag. This involves calculating the transition probabilities for abstract labels and the emission probabilities for entity types. However, the effects of these approaches are disturbed due to noisy non-entity tokens (i.e. "O" entity type) and inaccurate token-wise label dependency~\cite{08},~\cite{11} (see the example in Figure~\ref{img:label_dependency}).

Pipeline systems~\cite{04},~\cite{11}, ~\cite{53} detect entity spans in text and classify them into entity types. Current span detectors depend heavily on labor-intensive manual labeling for training guidance~\cite{46}. Most span detectors initially train on fundamental span features and then adapt to task-specific features~\cite{48}. This approach results in the repetitive training of fundamental span features across multiple span detectors. However, the lack of publicly available span detector checkpoints means that each user must train their own span detectors starting with learning basic span features. This process requires numerous expert data annotations and substantial computational resources for the \textbf{repetitive training} of basic span features.

\begin{figure}[th!]
\includegraphics[width=0.45\textwidth]{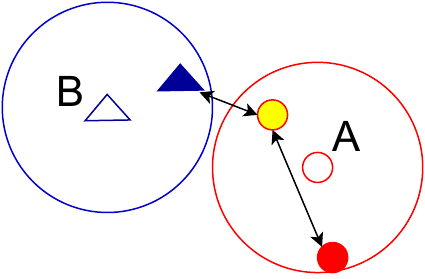}
\centering
\caption{A sample dependency problem in the 1-shot case. The yellow circle represents a query sample, while the red circle and purple triangle are 1-shot samples from classes A and B, respectively. Despite belonging to class A, the query sample (yellow circle) is misclassified as class B due to the absence of a closer reference sample from its own class. Instead, the nearest available reference is a purple triangle from class B, causing the query sample's misclassification. This misclassification stems from its sample dependency on the purple sample from class B.}
\label{img:sample_show}
\end{figure}

Both end-to-end and pipeline models often employ metric-based algorithms, such as the most widely used method, Prototypical Network~\cite{07},~\cite{09},~\cite{10}, and typically
use a specific metric that gauges the distance between the query sample and entity type referents,
ultimately assigning the most probable entity type to each token or span~\cite{49}. However, the limited number of samples for each type may not sufficiently represent the type referent, leading to heavy reliance on few-shot samples and causing the \textbf{sample dependency problem (SDP)}, which negatively affects the classifier performance (See Figure~\ref{img:sample_show}).

\begin{figure*}[ht!]
\includegraphics[width=0.9\textwidth,]{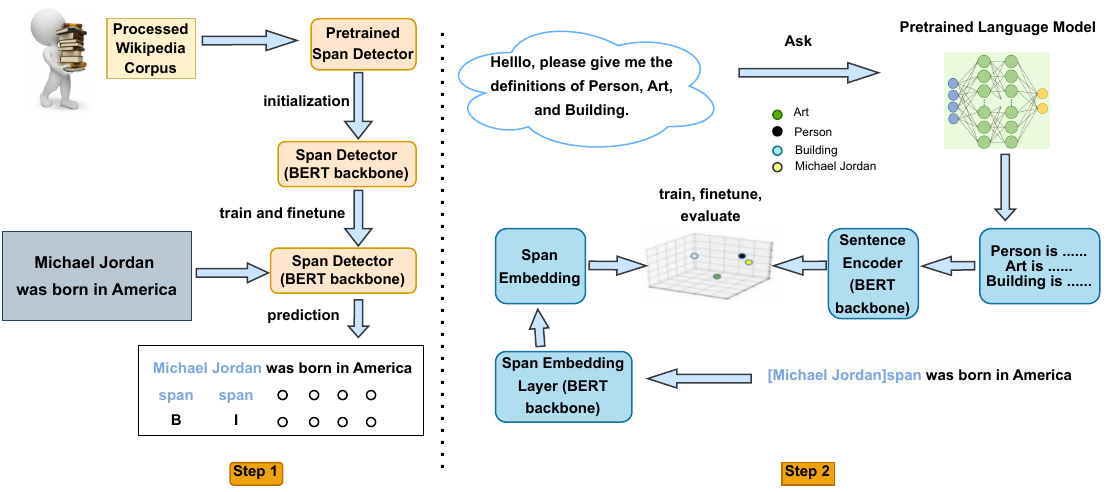}
\centering
\caption{SMCS model framework. In step 1, the Span Detector is initialized with a Pre-trained Span Detector and learns in the source domain. It then finetunes on the target domain to detect spans. In step 2, a pre-trained language model like GPT-3.5 generates entity type definitions, encoding them as \textbf{type referents} in a vector space. The predicted span's embedding from step 1 is compared to the type referents to determine its entity type.}
\label{fig:model_framework}
\end{figure*}

In this study, we address these problems and propose a more effective few-shot NER (fewNER) pipeline (Figure~\ref{fig:model_framework}). We trained a span detector on Wikipedia web data as a \textbf{steppingstone} span detector, which can be used to initialize the pipeline span detector to reduce the repetitive training of the basic span features and enable faster convergence when adapting to different domains. This approach saves machine resources and efforts for dataset annotation (details in the model section). To overcome the sample dependency problem in existing classifiers, we generate entity type definitions by leveraging \textbf{machine common sense} from large language models, such as GPT-3.5~\cite{12}. These definitions are encoded into the vector space as type referents along with the span representations for the similarity search. Our model, referred to as the \textbf{SMCS}~\footnote{To facilitate reading, all abbreviations used in this paper are compiled in the Acronyms Table located in the Appendix.}, initializes with a \textbf{s}teppingstone span detector and incorporates \textbf{m}achine \textbf{c}ommon \textbf{s}ense into an entity classifier.

Our study has a significant practical value for real-world applications. This publicly released steppingstone span detector can serve as an effective initialization tool for future span detectors. This aids in saving computational resources in research and industrial sectors. Furthermore, many companies require fine-grained named entity recognition models to extract detailed information from their vast document repositories~\cite{50}. However, owing to privacy concerns and sensitive information in document repositories, large language models may not be applicable~\cite{51}. Moreover, annotating training data for fine-grained NER is expensive because of  fine-grained expert knowledge, so only a few annotated samples are available~\cite{52}. Our study presents a methodology for constructing entity-type referents for such scenarios. Overall, our study provides practical solutions for real-world applications.

\begin{figure}[t]
\includegraphics[width=0.45\textwidth]{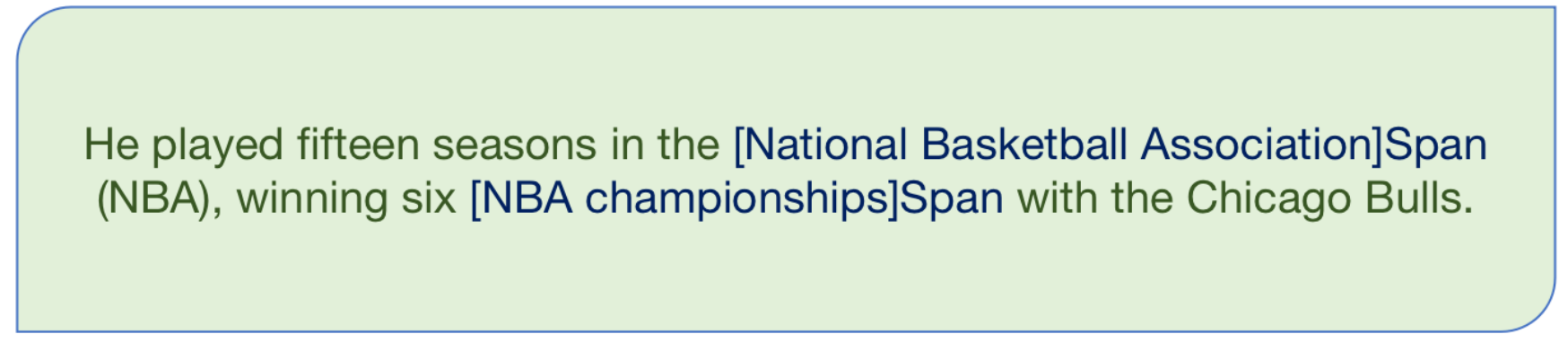}
\centering
\caption{Automatic annotation of Wikipedia web data.}
\label{img:spanannotation}
\end{figure}

Our contributions are as follows~\footnote{The project resources are in the GitHub \url{https://github.com/changtianluckyforever/SMCS_project}.}:
\begin{itemize}
\item (\textbf{a}) We released our steppingstone span detector with basic features, facilitating its integration into future span detectors to reduce repetitive training. In addition, we offer a substantial span detection dataset (Figure~\ref{img:spanannotation}) sourced from Wikipedia texts and labeled it automatically without consuming human resources.
\item (\textbf{b}) In this study, we identify and address the sample dependency problem in metric-based few-shot NER methods by leveraging machine common sense to construct entity-type referents. The empirical results demonstrate that our SMCS model outperforms other metric-based baselines on few-shot benchmark datasets, thereby demonstrating the effectiveness of our method in mitigating the sample dependency.
\item (\textbf{c}) SMCS leverages machine common sense in the entity-type classifier. This is the \textbf{first} study to utilize machine common sense to establish entity-type referents in few-shot NER.
\end{itemize}


\section{Related work}
Based on the model structure, few-shot NER models are typically categorized into two types~\cite{13}: pipeline structures and end-to-end structures. Different from existing methods, our approach is the first to apply machine common sense in forming class centroids to reduce sample dependency. We also provide a new module and dataset for community use.

\textbf{End-to-end Few-shot NER}. Previous works~\cite{07},~\cite{14},~\cite{15} employed token-level prototypical networks but relied on strong label dependencies for optimal performance~\cite{16}. However, accurately transferring abstract label distributions from the source to the target domains remains challenging in few-shot scenarios. \cite{16} attempted to replicate estimated label distributions from source to target but violated the probability definitions. \cite{08} addressed this by equally distributing source-domain probabilities in the target domain, but this contradicted real-world distributions.~\cite{09} introduced contrastive learning while preserving the label distribution scheme of ~\cite{08}, and~\cite{10} demonstrated the adverse effects of roughly estimated distributions on model performance. 
TLDT modifies optimization-based meta-learning by focusing on label dependency initialization and update rules~\cite{38}.~\cite{40} proposes a prototype retrieval and bootstrapping algorithm to identify representative clusters for each fine class, coupled with a mixture prototype loss learning class representations. The work~\cite{42} intervenes in the context to block the backdoor path between context and label. Prompt-based models have been studied extensively~\cite{06},~\cite{17},~\cite{18}. However, the performance of prompt-based methods~\cite{18} relies heavily on the chosen prompt~\cite{09}.
~\cite{43} employs instruction finetuning to comprehend and process task-specific instructions, including both main and auxiliary tasks. The PRM approach relies on human-generated descriptions that necessitate expert input and effort, proving less adaptable to new classes~\cite{37}. ~\cite{44} extracts entity type-related features based on mutual information criteria and generates a unique prompt for each unseen example by selecting relevant entity type-related features. ~\cite{46} utilizes external knowledge to establish semantic anchors for each entity type, appending these anchors to input sentence embeddings as template-free prompts for in-context optimization. And, the work~\cite{47} uses an image caption model to transform images into text descriptions, then ranks the nearest examples based on combined text and image similarity scores to create a demonstration context for in-context learning in NER. Differently, we used automatically generated definitions instead of handcrafted, resource-intensive prompts.

\textbf{Pipeline Few-shot NER}. ~\cite{11} decomposed the NER task into span detection and entity-type classification. However, their span detector often results in overlapping entity spans. Similarly, ~\cite{04} employed a decomposed pipeline with a deterministic span detector and used meta-learning to enhance the parameter initialization. In contrast, EP-Net~\cite{10} proposed prototypes with dispersed distributions that lacked semantic meaning. 
MANNER incorporates a memory module to store source domain information and uses optimal transport to effectively adapt this information to target domains~\cite{39}. TadNER~\cite{41} eliminates false spans by filtering out those not closely related to type names and develops more precise and stable prototypes using both support samples and type names. ~\cite{45} explicitly incorporates taxonomic hierarchy information into prototype representations and bridges the representation gaps between entity spans and taxonomic description embeddings. Our work differs by introducing a pre-trained span detector as a resource-saving steppingstone. Additionally, we pioneered the use of machine common sense to construct type referents, effectively mitigating the sample dependency problem encountered in previous studies.

\section{Task Formulation}
\label{sec:taskformulation}
Given a text sequence $\boldsymbol{X}=\left\{x_i\right\}_{i=1}^L$ as the input, an NER model is expected to output a label sequence $\boldsymbol{Y}=\left\{y_i\right\}_{i=1}^L$, where $L$ is the sequence length, ${x_i}$ is the $i$-th token, and ${y_i}$ is the entity label of token ${x_i}$. Usually, $y_i \in \boldsymbol{E}$, where $\boldsymbol{E}$ is a predefined set of entity classes, including the non-entity label $O$. Considering CoNLL-2003 as an example, where $\boldsymbol{E}$ denotes the set of entity classes: \{O, LOC, MISC, ORG, PER\}. 
\begin{itemize}
\item The label O indicates that the token does not belong to any specific entity class.
\item LOC signifies the token belongs to the location entity class.
\item ORG refers to the organization entity class.
\item PER indicates the token belongs to the person entity class.
\item MISC represents miscellaneous entities that do not fall into the previous three categories PER, ORG, and LOC.
\end{itemize}

In this study, we used the standard $N$-way $K$-shot (see Figure~\ref{img:22sample}) \textbf{episode-based few-shot setting} following existing works~\cite{09},~\cite{19},~\cite{20}. Each episode is a sample of the dataset that incorporates a support set, a query set, and a set of entity classes. The support set is N-way K-shot.

During the training stage, the training episodes ${C}_{\text {train }}=\left\{{S_{\text {train }}, Q_{\text {train }}, Y_{\text {train }}}\right\}$ are sampled from the source-domain labeled data. Here, the $m$-th episode ${C}^{m}_{\text {train }}=\left\{{S_{\text {train }}^{m}, Q_{\text {train}}^{m}, Y_{\text {train}}^{m}}\right\}$ is a sample of a batch of training data, 
where $S^{m}_{\text {train }}=\left\{\left(\boldsymbol{X}^{(i)}, \boldsymbol{Y}^{(i)}\right)\right\}_{i=1}^{N \times K}$ represents the $N$-way $K$-shot support set, $Q^{m}_{\text {train }}=\left\{\left(\boldsymbol{X}^{(j)}, \boldsymbol{Y}^{(j)}\right)\right\}_{j=1}^{N \times K^{\prime}}$ represents the corresponding query set containing $K^{\prime}$ shots for each of the $N$ entity classes, and $Y^{m}_{\text {train }}$ represents the set of entity classes (the cardinality of $Y^{m}_{\text {train }}$ is $N$). Similar to the training stage, in the testing stage, the testing episodes ${C}_{\text {test }}=\left\{{S_{\text {test }}, Q_{\text {test }}, Y_{\text {test }}}\right\}$ are sampled from the target-domain data in the same format. Here, an episode ${C}^{m}_{\text {test}}$ is a sample of a batch of test data. In the few-shot NER task, an NER model trained on training episodes ${C}_{\text {train}}$ will be fine-tuned on the $N$-way $K$-shot support set $S_{\text {test}}^{m}=\left\{\left(\boldsymbol{X}^{(i)}, \boldsymbol{Y}^{(i)}\right)\right\}_{i=1}^{N \times K}$ of the $m$-th episode $\left\{{S_{\text {test }}^{m}, Q_{\text {test }}^{m}, Y_{\text {test }}^{m}}\right\}$ from ${C}_{\text {test}}$, and then the NER model will be used to make predictions on the paired query set $Q_{\text {test}}^{m}=\left\{\boldsymbol{X}^{(j)}, \boldsymbol{Y}^{(j)}\right\}_{j=1}^{N \times K^{\prime}}$. $Y_{\text {test }}^{m}$ denotes the set of entity classes of $N$ types. In the few-shot setting, for any entity class set $Y_{\text {train}}$ from the source domain and entity class set $Y_{\text {test}}$ from the target domain, the condition $Y_{\text {train}} \cap Y_{\text {test}}=\emptyset$ holds.

\begin{figure*}[th!]
\includegraphics[width=0.95\textwidth]{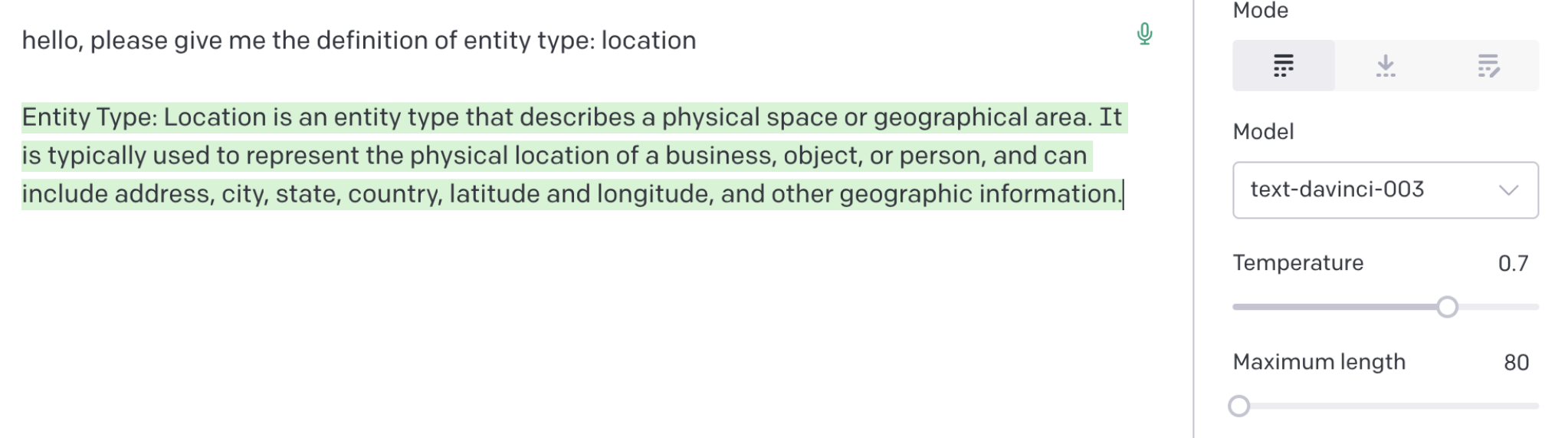}
\centering
\caption{An automatic definition for entity type: `location'. We ask GPT-3.5 to generate an automatic definition for the `location' entity type, which will then be encoded for the referent to classify relevant entities. We set the temperature to 0.7 and the maximum length to 80 and use the text-davinci-003 version of GPT-3.5.}
\label{fig:gpt3mcs}
\end{figure*}

\section{Model}
Our model represents an improved pipeline designed for few-shot NER. Many current span detectors require training from scratch to grasp fundamental span features, resulting in the repetitive training of fundamental span features across research and industry sectors. To address this, we have publicly released a foundational steppingstone span detector that encompasses these basic span features, enabling its use in initializing task-specific span detectors. Additionally, our model leverages machine common sense to construct entity-type referents, effectively mitigating the sample dependency problem.
\subsection{Entity Span Detector}
We utilized open-domain Wikipedia texts to create an automatically annotated dataset for entity spans. Subsequently, we trained our Steppingstone Span Detector by using this dataset. This detector can be employed to initialize span detectors in various domains, thereby effectively reducing the requirement for repetitive training.
\subsubsection{Steppingstone Span Detector}
\label{sec:universalspancorpus}
\begin{figure}[th!]
\includegraphics[width=0.45\textwidth,]{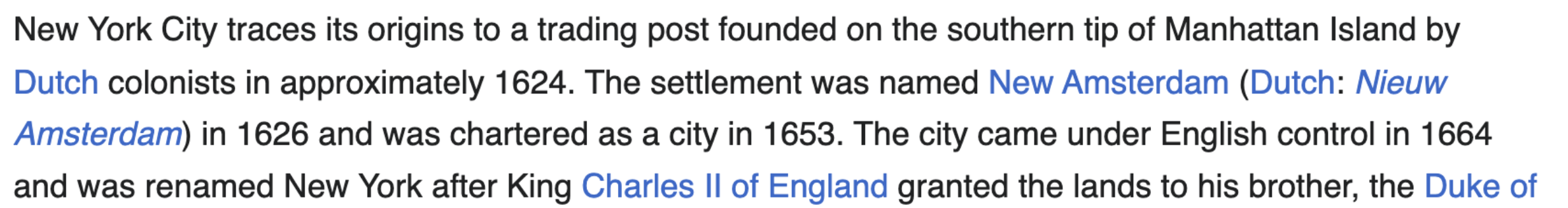}
\centering
\caption{New York introduction text snippet from the Wikipedia website.}
\label{img:new_york}
\end{figure}

\begin{figure}[ht!]
\includegraphics[width=\columnwidth,]{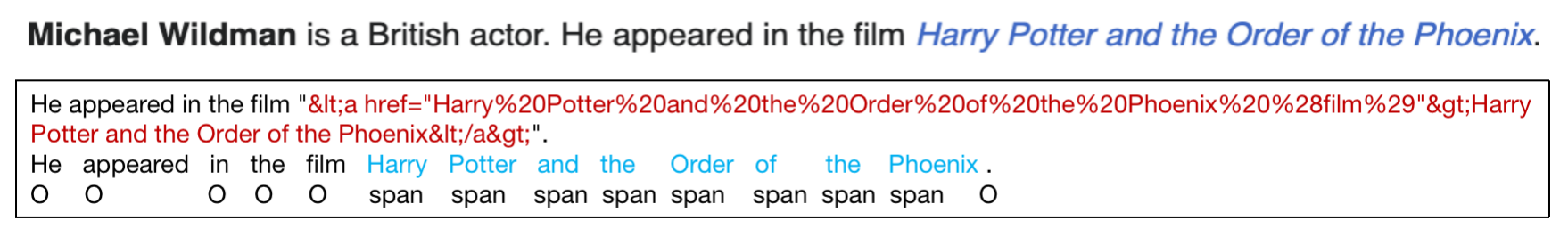}
\centering
\caption{An example of automatically annotating Wikipedia text spans.}
\label{img:span_hyperlinks}
\end{figure}

\textbf{Wikipedia Entity Span Dataset}~\footnote{This dataset will be publicly available upon acceptance.}: We leveraged Wikipedia's open-domain texts and automatically annotated entity spans, by using hyperlink-marked texts (Figure~\ref{img:new_york} and Figure~\ref{img:span_hyperlinks}). This process is performed programmatically without the need for human resources, thus allowing scalability. Our entity-span dataset comprises 52,400 sentences, each containing at least one entity span.

\textbf{Training of Span Detector}: We used the Wikipedia span dataset to train the steppingstone span detector via sequence labeling. For an input sequence $\boldsymbol{X}=\left\{x_i\right\}_{i=1}^L$ of length $L$, $\boldsymbol{T}=\left\{t_i\right\}_{i=1}^L$ is the associated label sequence. In this study, we used the BIOES scheme~\cite{21} because of its detailed annotation. We used the BERT-base-uncased module~\cite{22} as the text encoder, denoted by $f_\omega$. To facilitate convenient incorporation into future work, we stacked a logistic regression layer on top as the classifier. Taking a sequence $\boldsymbol{X}$ as input, it produces $p\left(x_i\right) \in \mathbb{R}^{|Label|}$ for each token $x_i$, with $Label=\{\textbf{B}, \textbf{I}, \textbf{O}, \textbf{E}, \textbf{S}\}$~\footnote{B means the start of the span, I is the inside of the span, O is the outside of the span, E represents the span end, and S represents a single-element span.} being the label set. We used the cross-entropy loss averaged over all tokens of sequence $\boldsymbol{X}$ as its loss: 
\begin{equation}
\label{eqn:loss_span_steppingstone}
\mathcal{L}=\frac{1}{L} \sum_{i=1}^L CrossEntropy \left(t_i, p\left(x_i\right)\right).
\end{equation}

\subsubsection{Domain Adaptation of Span Detector}
To improve domain adaptation and ensure a fair comparison, we use the MAML algorithm~\cite{23},~\cite{24} to search for the optimal model parameters as in previous studies~\cite{04},~\cite{24},~\cite{25},~\cite{26}.

\textbf{Training}.
In Step 1 of Figure~\ref{fig:model_framework}, we loaded our pre-trained steppingstone span detector to initialize the span detector with parameters $\pi$. 

Given a batch of training episodes $\left\{{C}^{m}_{\text {train }}\right\}^{B}_{m=1}$, and B is the batch size. Randomly sample a text sequence $\boldsymbol{X}$ from $\left\{{C}^{m}_{\text {train }}\right\}^{B}_{m=1}$, where $\boldsymbol{X}=\left\{x_i\right\}_{i=1}^L$ with length $L$, and $\boldsymbol{T}=\left\{t_i\right\}_{i=1}^L$ represents the associated label sequence.

We used the BERT-base-uncased module~\cite{22} as the text encoder  $f_\omega$, the encoder 
 $f_\omega$ produces contextualized representations $\boldsymbol{H}=\left\{h_i\right\}_{i=1}^L$ for a given sequence $\boldsymbol{X}$. We calculated the probability $p\left(x_i\right)$ that a token $x_i$ belongs to a span label, using $p\left(x_i\right)=\operatorname{softmax}\left(W h_i+b\right)$, where $p\left(x_i\right) \in \mathbb{R}^{|Label|}$ and $Label=\{\textbf{B}, \textbf{I}, \textbf{O}, \textbf{E}, \textbf{S}\}$ is the label set. The trainable parameter set of the span detector is denoted by $\pi=\{\omega, W, b\}$. We computed the training loss for the span detector as the average cross-entropy between the predicted and ground-truth label distributions for all tokens, as well as the maximum loss term for tokens with insufficient training (the second part in Equation~\ref{eqn:loss_span}). This results in the following cross-entropy loss for a text sequence $\boldsymbol{X}$:
\begin{equation}
\label{eqn:loss_span}
\begin{aligned}
\mathcal{L}(\pi) & =\frac{1}{L} \sum_{i=1}^L \text { CrossEntropy }\left(t_i, p\left(x_i\right)\right) \\
& +\zeta \max _{i \in\{1,2, \ldots, L\}} \text { CrossEntropy }\left(t_i, p\left(x_i\right)\right),
\end{aligned}
\end{equation}
where $\zeta$ is a coefficient used to adjust the weight of the maximum loss term.

We randomly sample a training episode in the batch, ${C}^{m}_{\text {train }}=\left\{{S_{\text {train }}^{m}, Q_{\text {train}}^{m}, Y_{\text {train}}^{m}}\right\}$, $m$ means the $m$-th training episode. $S_{\text {train }}^{m}$ contains multiple text sequences, and we derive the new model parameter set $\pi^{+}_{m}$ via $n$ steps of the inner update using the loss in Equation~\ref{eqn:loss_span} and computing on $S_{\text {train }}^{m}$:
\begin{equation}
\label{eqn:innerupdate}
\pi^{+}_{m} \leftarrow [\pi-\alpha 
   \nabla_{\pi} \mathcal{L}\left(\pi ;
S_{\text {train }}^{m} \right)]_{n  
 -steps} ,
\end{equation}
where $\alpha$ is the learning rate for the inner update, $\nabla_\pi$ means the gradient with regard to $\pi$, and $S_{\text {train}}^{m}$ contains multiple text sequences.

We then evaluated the updated parameter set $\pi^{+}_{m}$ on the query set $Q_{\text {train}}^{m}$ and further updated the span detector parameter set by minimizing the loss $\mathcal{L}\left(\pi^{+}_{m} ;
Q_{\text {train }}^{m} \right)$ with regard to $\pi$, which is the meta-update step of meta-learning. When aggregating all training episodes in the training batch, the meta-update objective is:
\begin{equation}
\label{eqn:second_order}
\min _{\pi}  \sum_{m=1}^{B} 
\mathcal{L}\left(\pi^{+}_{m} ;
Q_{\text {train }}^{m} \right),
\end{equation}
where $Q_{\text {train}}^{m}$ is the query set of the $m$-th training episode in the training batch and $Q_{\text {train}}^{m}$ contains multiple text sequences. B represents the number of episodes in the training batch.

Equation~\ref{eqn:second_order} involves a second-order derivative with regard to $\pi$, following a suggestion from previous studies~\cite{04},~\cite{23},~\cite{25}. We also employed a first-order approximation for
computational efficiency:
\begin{equation}
\pi^{*} \leftarrow \pi-\beta  \sum_{m=1}^{B} \nabla_{\pi^{+}_{m}} \mathcal{L}\left(\pi^{+}_{m} ; 
Q_{\text {train}}^{m} \right),
\end{equation}
where $\beta$ is the learning rate of the meta-update process. After training, $\pi^{*}$ is the span detector parameter set for further domain adaptation.

\textbf{Testing}. During testing, we fine-tune the meta-updated span detector $\pi^{*}$ on the support set $S_{\text {test}}^{m}$ from the test episode ${C}^{m}_{\text {test }}=\left\{{S_{\text {test}}^{m}, Q_{\text {test}}^{m}, Y_{\text {test}}^{m}}\right\}$ using the loss function defined in Equation~\ref{eqn:innerupdate} to obtain $\pi^{t}$. In the $N$-way $K$-shot setting, each target entity type $y_n \in Y_{\text {test}}^{m}$ has $K$ support samples from support set $S_{\text {test}}^{m}$ (as illustrated in Figure~\ref{img:22sample}). We then evaluated the span detection performance of $\pi^{t}$ on the associated query set. 

During the evaluation, we predicted the label distribution with $\pi^{t}$ for each token in the unseen sentences of $Q_{\text {test}}^{m}$, and decode using the Viterbi algorithm~\cite{27}. Constraints were enforced in the transition matrix to ensure adherence to the selected BIOES labeling scheme. In the BIOES labeling scheme, certain label transitions, such as from B to S, I to B, S to I, etc., are not permissible. To address this, we assigned an extremely small probability to these transitions to prevent their occurrence. However, other transitions under the BIOES labeling scheme are not subject to such constraints.

\subsection{Entity Classification}
In few-shot NER tasks, many metric-based entity classifiers rely on a limited number of available samples to establish entity-type referents for the classification. This often leads to the sample dependency problem (see Figure~\ref{img:sample_show}) within few-shot NER tasks. In contrast, our entity classifier employs machine common sense to construct entity type referents. This approach helps to mitigate the sample dependency problem.
\subsubsection{Machine Common Sense}
Common sense~\cite{28} is a universally shared practical knowledge derived from evidence and experience. Although large language models (LLMs), such as GPT-3.5~\cite{29} capture some common sense, few-shot NER faces challenges in capturing the semantics and common sense of entity classes. To overcome this, we propose using LLMs to automatically generate entity-type definitions for constructing general type referents to address the sample dependency problem in few-shot scenarios. We utilize the entity types Location and Date as illustrative examples (see Table~\ref{tab:mcs_type_definition}).
We specifically chose GPT-3.5 and encoded the generated definitions using BERT-base-uncased~\cite{22} for a fair comparison with previous work in few-shot NER.
\begin{table}[t]
\resizebox{\columnwidth}{!}{%
\begin{tabular}{ll}
\hline
 &
  Machine Common Sense \\ \hline
Location &
  \begin{tabular}[c]{@{}l@{}}Location is an entity type that describes a physical \\ space or geographical area. It is typically used to \\ represent the physical location of a business, object, \\ or person, and can include address, city, state, \\ country,  latitude and longitude, and other \\ geographic information.\end{tabular} \\ \hline
Date &
  \begin{tabular}[c]{@{}l@{}}Date is a type of named entity that refers to \\ a specific point or period of time. It can be expressed \\ using a variety of formats, such as a calendar date, \\ an ordinal date,  or even a relative date.\end{tabular} \\ \hline
\end{tabular}%
}
\caption{Examples of machine common sense for entity type definitions.}
\label{tab:mcs_type_definition}
\end{table}

\subsubsection{Entity Span Classification}
We utilized MAML, a meta-learning algorithm~\cite{23}, for parameter initialization to improve domain adaptation.

\textbf{Training}.
In Step 2 of Figure~\ref{fig:model_framework}, the automatically generated entity-type definitions from GPT-3.5 are encoded by the BERT-base-uncased~\cite{22} sentence encoder $f_\tau$ into entity-type referents. The entity classifier parameter set $\mu$ includes $\kappa$, $\tau$, and $W_c$, which represent the entity span encoder, sentence encoder, and classification module, respectively. $\mu=\{\kappa, \tau, W_c\}$. We use the BERT-base-uncased~\cite{22} model for all encoders, to ensure fair comparisons with the baselines in the experiments.

Given a batch of training episodes $\left\{{C}^{m}_{\text {train }}\right\}^{B}_{m=1}$, and B is the batch size. Randomly sampling a text sequence $\boldsymbol{X}$ from $\left\{{C}^{m}_{\text {train }}\right\}^{B}_{m=1}$, where $\boldsymbol{X}=\left\{x_i\right\}_{i=1}^L$ with length $L$, the span encoder $f_\kappa$ generates contextualized representations $\boldsymbol{H}=f_\kappa(\boldsymbol{X})=\left\{h_i\right\}_{i=1}^L$.

The entity span of sequence $\boldsymbol{X}$ which starts at $x_i$ and ends at $x_j$ is denoted by $x_{[i, j]}$. We computed the span representation of $x_{[i, j]}$ by averaging
all the token representations in $x_{[i, j]}$:
\begin{equation}
s_{[i, j]}=\frac{1}{j-i+1} \sum_{k=i}^j h_k.
\end{equation}

For each potential entity class $y_n$ for the entity span $x_{[i, j]}$, we used the large language model GPT-3.5 (text-davinci-003) to generate the definition $\boldsymbol{P}^{n}$ (see Figure~\ref{fig:gpt3mcs} for details), $\boldsymbol{P}^{n}$ is a text sequence with length \textit{J}. 
BERT-base-uncased~\cite{22} sentence encoder $f_\tau$ generates contextualized representations: $\boldsymbol{H}^{n}=f_\kappa(\boldsymbol{P}^{n})=\left\{h_i^{n}\right\}_{i=1}^J$. Then, we average all the token embeddings of $\boldsymbol{P}^{n}$ to compute the entity-type referent embedding $V^n$:
\begin{equation}
\label{eqn:promptvector}
V^n=\frac{1}{J} \sum_{i=1}^J h_i^{n}.
\end{equation}
The probability that entity span $x_{[i, j]}$ belongs to each entity type $y_n$ is:
\begin{equation}
\label{eqn:type_infer}
p\left(y_n | x_{[i, j]}\right)=\operatorname{sigmoid}\left(W_c(s_{[i, j]}, V^n,|s_{[i, j]}-V^n|)\right), 
\end{equation}
where $|s_{[i, j]}-V^n|$ is the absolute difference between the two vectors, $s_{[i, j]}, V^n, |s_{[i, j]}-V^n|$ are concatenated as a new vector for $W_c$.

By computing on the text sequence $\boldsymbol{X}$, the parameter set $\mu$ of the entity classifier is updated using the following cross-entropy loss function:
\begin{equation}
\label{eqn:loss_type}
\begin{aligned}
\mathcal{L}(\mu) & = \frac{1}{N}   \sum_{x_{[i, j]} \in \boldsymbol{X} }    \text { CrossEntropy }\left(y_{[i, j]}, x_{[i, j]}\right) \\
& +\zeta \max _{x_{[i, j]} \in \boldsymbol{X} } \text { CrossEntropy }\left(y_{[i, j]}, x_{[i, j]}\right),
\end{aligned}
\end{equation}
where $\zeta$ is the weight of the maximum span classification loss term, $y_{[i, j]}$ denotes the ground-truth entity class corresponding to span $x_{[i, j]}$, and $\mu=\{\kappa, \tau, W_c\}$ are the trainable parameters. N is the number of entity spans within text sequence $\boldsymbol{X}$.

We randomly sampled a training episode in the batch; ${C}^{m}_{\text {train }}=\left\{{S_{\text {train }}^{m}, Q_{\text {train}}^{m}, Y_{\text {train}}^{m}}\right\}$, $m$ means the $m$-th training episode. $S_{\text {train }}^{m}$ contains multiple text sequences, and we derive the new model parameter set $\mu^{+}_{m}$ via $n$ steps of the inner update using the loss in Equation~\ref{eqn:loss_type} and computing on $S_{\text {train }}^{m}$:
\begin{equation}
\label{eqn:entitynsteps}
\mu^{+}_{m} \leftarrow [\mu-\alpha 
   \nabla_{\mu} \mathcal{L}\left(\mu ;
S_{\text {train }}^{m} \right)]_{n  
 -steps} ,
\end{equation}
where $\alpha$ is the learning rate for the inner update, $\nabla_\mu$ means the gradient with regard to $\mu$, and $S_{\text {train}}^{m}$ contains multiple text sequences.

We then evaluated the updated parameter set $\mu^{+}_{m}$ on the query set $Q_{\text {train}}^{m}$ and further updated the entity classifier parameter set by minimizing the loss $\mathcal{L}\left(\mu^{+}_{m} ;
Q_{\text {train }}^{m} \right)$ with regard to $\mu$, which is the meta-update step of meta-learning. When aggregating all training episodes in the training batch, the meta-update objective is:
\begin{equation}
\label{eqn:type_order}
\min _{\mu}  \sum_{m=1}^{B} 
\mathcal{L}\left(\mu^{+}_{m} ;
Q_{\text {train }}^{m} \right),
\end{equation}
where $Q_{\text {train}}^{m}$ is the query set of the $m$-th training episode in the training batch and $Q_{\text {train}}^{m}$ contains multiple text sequences. B is the number of episodes in the training batch.

Because Equation~\ref{eqn:type_order} involves a second-order derivative with regard to $\mu$, following a suggestion from previous studies~\cite{04},~\cite{23},~\cite{25}, we also employ its first-order approximation for
computational efficiency:
\begin{equation}
\mu^{*} \leftarrow \mu-\beta  \sum_{m=1}^{B} \nabla_{\mu^{+}_{m}} \mathcal{L}\left(\mu^{+}_{m} ; 
Q_{\text {train}}^{m} \right),
\end{equation}
where $\beta$ is the learning rate of the meta-update process. After training, $\mu^{*}$ is the entity classifier parameter set for further domain adaptation.

\textbf{Test}. During the testing stage, we fine-tuned the meta-update entity classifier $\mu^{*}$ using the loss function in Equation~\ref{eqn:entitynsteps} on the support set $S_{\text {test}}^{m}$ from the test episode ${C}^{m}_{\text {test }}=\left\{{S_{\text {test}}^{m}, Q_{\text {test}}^{m}, Y_{\text {test}}^{m}}\right\}$, resulting in $\mu^{t}$. In the N-way K-shot setting, each target entity type $y_n \in Y_{\text {test}}^{m}$ has K support samples in $S_{\text {test}}^{m}$ (see Figure~\ref{img:22sample} for details). We evaluate the entity classifier's performance by inferring the entity type for each detected entity span $x_{[i, j]}$ in $Q_{\text {test}}^{m}$, using the label $y_n \in Y_{\text {test}}^{m}$ with the highest probability in Equation~\ref{eqn:type_infer}:

\begin{equation}
y_{[i, j]}=\arg \max _{y_n} p\left(y_n | x_{[i, j]}\right)
\end{equation}

\begin{table*}[t]
\resizebox{\textwidth}{!}{%
\begin{tabular}{rrrrrrrrr}
\hline
\multicolumn{1}{c}{\multirow{2}{*}{\textbf{Models}}} & \multicolumn{4}{c}{\textbf{1-shot}}               & \multicolumn{4}{c}{\textbf{5-shot}}               \\ \cline{2-9} 
\multicolumn{1}{c}{}                                 & CoNLL      & GUM        & WNUT       & Ontonotes  & CoNLL      & GUM        & WNUT       & Ontonotes  \\ \hline
MatchingNetwork                                      & 19.50±0.35 & 4.73±0.16  & 17.23±2.75 & 15.06±1.61 & 19.85±0.74 & 5.58±0.23  & 6.61±1.75  & 8.08±0.47  \\
L-TapNET+CDT                                         & 44.30±3.15 & 12.04±0.65 & 20.80±1.06 & 15.17±1.25 & 45.35±2.67 & 11.65±2.34 & 23.30±2.80 & 20.95±2.81 \\
MAML-ProtoNet                                        & 46.09±0.44 & 17.54±0.98 & 25.14±0.24 & 34.13±0.92 & 58.18±0.87 & 31.36±0.91 & 31.02±1.28 & 45.55±0.90 \\
ChatGPT                                        & \textbf{47.57±3.16} & 14.05±1.97 & 26.24±3.41 & 35.76±2.87 & \textbf{61.90±4.36} & 18.01±3.92 & 27.50±2.76 & 40.25±4.13 \\
TLDT                                       & 43.05±1.70 & 5.74±0.79 & 27.94±1.42 & 22.07±1.06 & 49.31±4.99 & 10.14±1.60 & 29.08±2.70 & 31.03±3.07 \\
TFP                                       & 46.16±0.57 & 17.72±0.38 & 25.89±0.51 & 35.32±1.07 & 59.23±0.69 & 29.77±0.52 & 31.61±0.87 & 44.92±1.21 \\

SMCS(ours) &
  47.21±0.81 &
  \textbf{18.10±0.38} &
  \textbf{28.58±0.37} &
  \textbf{36.70±0.41} &
  59.76±0.85 &
  \textbf{31.92±0.61} &
  \textbf{33.15±0.67} &
  \textbf{48.32±0.74} \\ \hline
\end{tabular}%
}
\caption{The cross-domain experiment results are displayed as micro F1 scores with standard deviations. Averages are based on 5 runs. In 1-shot or 5-shot cases, each target entity type has 1 or 5 training samples, respectively. We use the results reported in the baseline papers. CoNLL dataset has 4 annotated entity classes, GUM dataset has 11 annotated entity classes, WNUT dataset has 6 annotated entity classes, and OntoNotes has 18 annotated entity classes.}
\label{tab:cross-domain-p}
\end{table*}

\begin{table*}[th!]
\resizebox{\textwidth}{!}{%
\begin{tabular}{llllllllll}
\hline
\multicolumn{1}{c}{\multirow{3}{*}{\textbf{Models}}} &
  \multicolumn{4}{c}{\textbf{Inter}} &
  \multicolumn{4}{c}{\textbf{Intra}} &
   \\
\multicolumn{1}{c}{} &
  \multicolumn{2}{c}{5 way} &
  \multicolumn{2}{c}{10 way} &
  \multicolumn{2}{c}{5 way} &
  \multicolumn{2}{c}{10 way} &
   \\
\multicolumn{1}{c}{} &
  \multicolumn{1}{c}{1 shot} &
  \multicolumn{1}{c}{5 shot} &
  \multicolumn{1}{c}{1 shot} &
  \multicolumn{1}{c}{5 shot} &
  \multicolumn{1}{c}{1 shot} &
  \multicolumn{1}{c}{5 shot} &
  \multicolumn{1}{c}{1 shot} &
  \multicolumn{1}{c}{5 shot} &
  Averaged \\ \hline
CONTAINER &
  55.95±xxx &
  61.83±xxx &
  48.35±xxx &
  57.12±xxx &
  40.43±xxx &
  53.70±xxx &
  33.84±xxx &
  47.49±xxx &
  49.84 \\
ESD &
  66.46±0.49 &
  \textbf{74.14±0.80} &
  59.95±0.69 &
  67.91±1.41 &
  41.44±1.16 &
  50.68±0.94 &
  32.29±1.10 &
  42.92±0.75 &
  54.47 \\
EP-Net &
  62.49±0.36 &
  65.24±0.64 &
  54.39±0.78 &
  62.37±1.27 &
  43.36±0.99 &
  58.85±1.12 &
  36.41±1.03 &
  46.40±0.87 &
  53.60 \\
MAML-ProtoNet &
  68.77±0.24 &
  71.62±0.16 &
  63.26±0.40 &
  68.32±0.10 &
  52.04±0.44 &
  63.23±0.45 &
  43.50±0.59 &
  56.84±0.14 &
  60.99 \\
COPNER &
  65.39±xxx &
  67.59±xxx &
  59.69±xxx &
  62.32±xxx &
  53.52±xxx &
  58.74±xxx &
  44.13±xxx &
  51.55±xxx &
  57.87 \\
TFP &
  70.83±0.62 &
  72.14±0.40 &
  \textbf{64.70±0.72} &
  67.65±0.15 &
  55.49±0.67 &
  \textbf{63.31±0.77} &
  \textbf{46.29±0.74} &
  54.01±0.60 &
  61.80 \\
2INER &
  \textbf{70.90±0.87} &
  71.28±1.21 &
  62.18±0.65 &
  63.72±0.83 &
  \textbf{56.02±0.71} &
  58.69±0.67 &
  45.57±0.82 &
  49.61±0.96 &
  59.75 \\
ChatGPT &
  36.88±3.98 &
  39.02±4.27 &
  30.84±3.61 &
  36.45±3.13 &
  30.77±2.89 &
  38.61±3.76 &
  24.78±2.71 &
  30.07±4.62 &
  33.43 \\
SMCS(ours) &
  68.19±0.63 &
  71.33±0.47 &
  63.89±0.39 &
  \textbf{68.81±0.23} &
  52.31±0.27 &
  63.03±0.16 &
  46.02±0.73 &
  \textbf{57.76±0.44} &
  61.42 \\ \hline
\end{tabular}%
}
\caption{In-domain Few-NERD experiment results are presented via micro F1 scores with standard deviations, spanning inter and intra settings. \textbf{Inter-5-1} signifies 5 target entity types with 1 training sample each, within the Inter setting. This abbreviation logic applies to comparable cases. Averages are based on 5 runs. We use the results reported in the baseline papers. The baseline papers, COPNER and CONTAINER, do not provide standard deviations for their original results. Few-NERD is a fine-grained NER dataset, which has 66 annotated entity classes.}
\label{tab:fewnerdperformance}
\end{table*}

\section{Experiment}
\begin{table}[th!]
\begin{tabular}{cccc}
\hline
\textbf{Dataset} & \textbf{Domain} & \textbf{Sentences} & \textbf{Classes} \\ \hline
Few-NERD~\cite{19}         & Mixed           & 188.2 k            & 66               \\
CoNLL03~\cite{30}          & News            & 20.7 k             & 4                \\
GUM~\cite{31}              & Wiki            & 3.5 k              & 11               \\
WNUT~\cite{32}             & Social          & 5.6 k              & 6                \\
OntoNotes~\cite{33}        & Mixed           & 159.6 k            & 18               \\ \hline
\end{tabular}
\centering
\caption{The statistics of datasets in the experiments.}
\label{tab:data_statistics}
\end{table}
\subsection{Experiment Setup}
\subsubsection{Datasets and Baselines}
We conducted experiments with two widely-used N-way K-shot based datasets: the in-domain dataset Few-NERD~\cite{19} and cross-domain dataset Cross-Dataset~\cite{16}. \textbf{Few-NERD} is annotated with a hierarchy of 8 coarse-grained and 66 fine-grained entity types. It has two tasks: 
\begin{itemize}
\item (a) \textbf{Inter}, where all entities in the train/dev/test splits may share coarse-grained types while keeping the fine-grained entity types mutually disjoint.
\item (b) \textbf{Intra}, where all entities in the train/dev/test splits belong to different coarse-grained types.
\end{itemize}
\textbf{Cross-Dataset} consists of 4 different NER domains:
\begin{itemize}
\item CoNLL-03 (news)~\cite{30},
\item OntoNotes 5.0 (mixed domain)~\cite{33},
\item WNUT-17 (social media)~\cite{32},
\item and GUM (Wikipedia)~\cite{31}.
\end{itemize}
The datasets statistics are presented in Table~\ref{tab:data_statistics}. In training, we selected two sets as the training set, and the remaining two others were used for the testing and development sets. To make a fair comparison, we utilized the processed episode data released by \cite{19} in Few-NERD and by~\cite{04} in the Cross-Dataset.

SMCS is a pipeline that uses machine common sense to construct type referents to mitigate the sample dependency problem; therefore, we choose recent state-of-the-art (SOTA) baselines that share common attributes with ours or belong to the same model category as ours. 

We compare with the following pipeline models: 
\begin{itemize}
\item ESD~\cite{11},
\item EP-Net~\cite{10},
\item MAML-ProtoNet~\cite{04}, We compare our SMCS method with MAML-ProtoNet due to the sample dependency problem in traditional ProtoNet, which uses sample averaging for class centroids. Applying machine common sense for centroid setting in SMCS, we aim to address this issue. Comparing with the basic ProtoNet version, as employed by MAML-ProtoNet, is crucial for validating our approach, as other versions with additional components may confound the evaluation.
\end{itemize}

We also compare with prompt-based models:
\begin{itemize}
\item COPNER~\cite{20}, which is a leading method among prompt-based non-large language models,
\item TFP~\cite{46}, utilizes external knowledge to establish semantic anchors for each entity type, appending these anchors to input sentence embeddings as template-free prompts for in-context optimization,
\item 2INER~\cite{43}, employs instruction finetuning to comprehend and process task-specific instructions, including both main and auxiliary tasks,
\item ChatGPT~\footnote{We use ChatGPT version gpt-3.5-turbo-1106 as a baseline in both the in-domain and cross-domain experiments.} on both in-domain and cross-domain benchmarks using an in-context learning setting. The prompts used for ChatGPT can be found in the Appendix.
\end{itemize}

For completeness, we also compare with representative end-to-end models:
\begin{itemize}
\item L-Tapnet+CDT~\cite{16},
\item TLDT~\cite{38},
\item Matching Network~\cite{34},
\item CONTAINER~\cite{09}.
\end{itemize}
Please refer to the Appendix for all baselines and prompt details. The prompts used for ChatGPT can be found in the Appendix.
\subsubsection{Evaluation Metrics and Implementation Details}
We followed the tradition in the previous few-shot NER tasks~\cite{04},~\cite{19},~\cite{20}, and used the \textbf{F1 score} as the evaluation metric. 
We also used the \textbf{convergence steps} as an evaluation metric, which represents the number of training steps for the models to converge. We defined the convergence status as a model that achieves good performance and is stable on the validation set during training. 

For each experiment, we performed five runs and averaged the results to obtain the final performance. We used a \textbf{grid search} to find the best hyper-parameters for each benchmark. The random seeds were 171, 354, 550, 667, and 985. We choose AdamW~\cite{35} as the optimizer with a warm-up rate of 0.1. Please refer to the Appendix for further details. 

\subsubsection{Research Questions}
We want to investigate: (a) overall results comparison on benchmarks, (b) quantitative analyses of SCMS modules under different settings, and (c) ablation study of entity type referents.

\subsection{Results}
As our SMCS model is not a large language model (LLM), this section initially focuses on comparing SMCS with non-LLM baselines. The comparison with the LLM, ChatGPT, is presented in the following section.

In Table~\ref{tab:cross-domain-p} cross-domain setting, SMCS excels in 1-shot and 5-shot tests due to its use of machine common sense, alleviating the sample dependency problem (SDP). 
SMCS outperforms TLDT and TFP, demonstrating the effectiveness of machine common sense in NER tasks. This is because TLDT, despite its focus on label dependency through initialization and update rules, falls short in its entity typing module. Meanwhile, TFP's use of external knowledge for semantic anchors and prompt construction for in-context learning shows potential, but its prompts require further refinement.
SMCS proportionally outperforms MAML-ProtoNet more in 1-shot than 5-shot cases, where metric-based methods benefit from superior type referents with more samples. This is further discussed in the Entity Classifier Analysis section.

Table~\ref{tab:fewnerdperformance} further examines models in N-way K-shot setups, considering (\textbf{1}) model comparison, (\textbf{2}) shot settings, (\textbf{3}) way settings, and (\textbf{4}) inter/intra scenarios.

(\textbf{1}) EP-Net~\cite{10} constructs new classification referents, but their semantic information is lacking, resulting in inferior results compared to ours. Among the prompt-based non-LLMs for few-shot NER, COPNER~\cite{20} is a leading method on the Few-NERD dataset; however, our model's use of automatic definitions generally outperforms it. We achieved similar results to MAML-ProtoNet under certain settings (Table~\ref{tab:fewnerdperformance}), with the fine-grained granularity discussed in the Entity Classifier Analysis section as the main factor. Overall, our SMCS model outperformed MAML-ProtoNet. 
SMCS matches TFP's performance under in-domain settings, suggesting that both prompt-learning and metric-based approaches are viable for NER. 2INER, using instruction finetuning for task-specific instructions, shows promising one-shot results. However, its prompts, favoring context over support samples with increased availability, need enhancement.
(\textbf{2}) In the 1-shot case, the SDP is more pronounced, and the SMCS has greater advantages over MAML-ProtoNet than in 5-shot cases due to the latter has more samples. Detailed explanations are provided in the subsequent sections. 
We observe that increasing support entity samples from 1 to 5 enhances all methods' performance, as more samples offer richer information. This highlights that limited support examples are the primary challenge in few-shot learning.
(\textbf{3}) As the number of classes (ways) increases from 5 to 10 (Table~\ref{tab:fewnerdperformance}), SMCS distinguishes itself from baselines by providing more accurate type referents and maintaining granularity rather than relying on limited samples for numerous challenging categories. 
(\textbf{4}) In Inter settings, test sets share outer layer types with train sets, unlike in Intra settings. Intra settings disperse the ground-truth referents, magnifying the sample dependency problem for metric-based methods (MAML-ProtoNet vs. SMCS). 
However, few-shot learning rarely has hierarchical annotation in reality (inter settings), whereas Intra setting remains frequent in reality. Our SMCS performs better than MAML-ProtoNet in the Intra settings.

\subsection{Comparison with ChatGPT}
\begin{table}[th!]
\resizebox{0.55\columnwidth}{!}{%
\begin{tabular}{llll}
\hline
Models                      & Model Size      \\ \hline
ChatGPT  & 175 Billion   \\
SMCS     & 220 Million  \\ \hline
\end{tabular}%
}
\centering
\caption{The number of model parameters comparison.}
\label{tab:parameters_number}
\end{table}
Drawing from the cross-domain results presented in Table~\ref{tab:cross-domain-p} and the in-domain results highlighted in Table~\ref{tab:fewnerdperformance}, our SMCS model consistently exhibited superior performance compared with ChatGPT across various scenarios. Notably, SMCS demonstrates significantly better performance on the FewNERD dataset than ChatGPT. However, on the CoNLL dataset, ChatGPT surpasses our model, SMCS, in both the 1-shot and 5-shot cases. The CoNLL NER dataset is coarse-grained with only 4 target classes, but SMCS still achieves results comparable to those of ChatGPT. 

Hence, our empirical observations suggest that while ChatGPT excels in coarse-grained settings, it exhibits weaker performance than SMCS in fine-grained contexts. Notably,  in practical applications, most NER problems are fine-grained and involve more than four classes.

Several factors might contribute to this observation. ChatGPT is pre-trained on an extensive corpus and boasts a model parameter volume of 175 billion (Table~\ref{tab:parameters_number}), which is significantly larger than the SMCS model size. This extensive knowledge repository aids ChatGPT in excelling coarse-grained NER tasks such as the CoNLL NER task. However, because its pre-training corpora are not fine-grained overall, this could explain why ChatGPT exhibits relatively poorer performance than non-LLMs in fine-grained NER tasks.
\subsection{Span Detector Analysis}
Our SMCS model operates as a pipeline comprising a span detector and entity classifier. To assess the individual influence of each module, we conducted quantitative performance analysis. MAML-ProtoNet, a significant baseline utilizing a pipeline structure, serves as a prominent reference point for both in-domain and cross-domain comparisons. Through quantitative analyses, we compared SMCS directly with MAML-ProtoNet, aiming to delineate the advantages of our model. This section focuses on the comparison of the span detector, while the subsequent section delves into a comparison of the entity classifier.

We compared the SMCS span detector with MAML-ProtoNet on Cross-Dataset and Few-NERD datasets. Using Steppingstone Span Detector (SSD) initialization, SMCS converges faster than MAML-ProtoNet (Table~\ref{tab:speedofconveregence}), reducing repetitive training. SSD serves a purpose akin to the role of BERT~\cite{22} in embeddings.
\begin{table}[th!]
\resizebox{\columnwidth}{!}{%
\begin{tabular}{lllll}
\hline
\multirow{2}{*}{Span detector}                                               & \multicolumn{2}{c}{Inter 5 way} & \multicolumn{1}{c}{CoNLL} & \multicolumn{1}{c}{GUM} \\ \cline{2-5} 
 & 1 shot & 5 shot & 1 shot & 5 shot \\ \hline
\begin{tabular}[c]{@{}l@{}}MAML-ProtoNet\end{tabular} & 668±144         & 364±89        & 264±68                    & 130±14                  \\
\begin{tabular}[c]{@{}l@{}}SMCS\end{tabular}          & 344±107         & 272±66        & 135±34                    & 90±11                   \\ \hline
\end{tabular}%
}
\caption{\textbf{Convergence steps} comparison between span detectors using 4 representative dataset settings, which contain in-domain datasets and cross-domain datasets respectively. We report the training steps before reaching the convergence status. Averages are based on 5 runs.}
\label{tab:speedofconveregence}
\end{table}

\begin{table}[th!]
\resizebox{\columnwidth}{!}{%
\begin{tabular}{llll}
\hline
Span detector                      & Inter-5-1  & GUM-1shot  & GUM-5shot  \\ \hline
MAML-ProtoNet  & 76.71±0.30 & 35.63±2.17 & 46.26±1.28 \\
SMCS           & 76.57+0.36 & 38.45±1.85 & 47.92±1.32 \\ \hline
\end{tabular}%
}
\caption{Span detector F1 scores comparison. We use 3 representative settings, which represent comparable results and better results respectively. Averages are based on 5 runs.}
\label{tab:spanf1scores}
\end{table}

Alongside faster convergence, we assessed span detector performance using F1 scores (Table~\ref{tab:spanf1scores}). As listed in Table~\ref{tab:spanf1scores}, our SMCS span detector outperformed the MAML-ProtoNet span detector on the GUM (wiki) dataset. Apart from the GUM dataset, the SMCS span detector performed similarly to the MAML-ProtoNet span detector on the other datasets, displaying minor performance differences similar to their performance comparison on the inter-5-1 dataset.

Therefore, in comparison with MAML-ProtoNet, for enhancements on the GUM dataset (Table~\ref{tab:cross-domain-p}), both the SMCS span detector and entity classifier play a role, whereas the entity classifier primarily drives improvements on other datasets (Tables~\ref{tab:cross-domain-p} and ~\ref{tab:fewnerdperformance}). Further details can be found in the Entity Classifier Analysis section.

\subsection{Entity Classifier Analysis}
\begin{table*}[th!]
\resizebox{\textwidth}{!}{%
\begin{tabular}{llllllllllllll}
\hline
\multicolumn{1}{c}{\multirow{2}{*}{Entity Classifier}} &
  \multicolumn{2}{c}{CoNLL} &
  \multicolumn{2}{c}{GUM} &
  \multicolumn{2}{c}{WNUT} &
  \multicolumn{2}{c}{Ontonotes} &
  \multicolumn{2}{c}{Intra-5way} &
  \multicolumn{1}{c}{Intra-10way} &
  \multicolumn{2}{c}{Inter-5way} \\
\multicolumn{1}{c}{} &
  1shot &
  5shot &
  1shot &
  5shot &
  1shot &
  5shot &
  1shot &
  5shot &
  1shot &
  5shot &
  1shot &
  1shot &
  5shot \\ \hline
MAML-ProtoNet &
  65.85±0.96 &
  76.14±0.75 &
  43.01±0.44 &
  56.40±1.33 &
  50.06±1.22 &
  61.88±1.13 &
  67.10±0.89 &
  78.05±0.45 &
  67.41±0.57 &
  80.30±0.35 &
  56.18±0.58 &
  \textbf{88.40±0.31} &
  93.39±0.18 \\
SMCS (ours) &
  \textbf{68.57±1.05} &
  \textbf{78.91±1.03} &
  \textbf{45.20±0.87} &
  \textbf{58.29±0.68} &
  \textbf{62.60±1.21} &
  \textbf{68.68±0.91} &
  \textbf{69.04±0.50} &
  \textbf{80.17±0.64} &
  \textbf{69.52±0.41} &
  \textbf{80.34±0.26} &
  \textbf{58.38±0.66} &
  88.01±0.47 &
  \textbf{93.57±0.21} \\ \hline
\end{tabular}%
}
\caption{Entity classifier comparison results using micro F1 scores on FewNERD and Cross-Dataset evaluations. Our analysis includes representative FewNERD settings: inter/intra-class, 5/10-way, 1/5-shot classifications. Our results include all sub-datasets of Cross-Dataset to cover cross-domain settings. Averages are based on 5 runs.}
\label{tab:entity_classifier_comparision}
\end{table*}
Our entity classifier~\footnote{We utilize the large language model solely for generating entity type definitions, without employing it for entity recognition and classification purposes for a fair comparison.} addresses the sample dependency problem (\textbf{SDP}) of metric-based classifiers, particularly the Prototypical Network, which is the most widely used. Therefore, we extensively compared our classifier with the MAML-ProtoNet classifier, and the results are presented in Table~\ref{tab:entity_classifier_comparision}. We analyzed them from three aspects: (\textbf{1}) Number of shots. (\textbf{2}) Specific cases. (\textbf{3}) Various datasets.

(\textbf{1}) The SMCS entity classifier outperformed MAML-ProtoNet in terms of the F1 score improvement. This is more proportionally prominent in 1-shot than 5-shot cases owing to biases from the limited samples. Metric-based methods rely heavily on samples and exhibit bias in 1-shot cases, thereby impacting NER classification. ProtoNet authors~\cite{36} admit bias in their type centroids owing to a few support samples, reaffirming this issue. MAML-ProtoNet improves in 5-shot cases with more proper type referents from more samples, yet bias remains. SMCS leverages machine common sense for better type referents and improves overall performance.

(\textbf{2}) In Table~\ref{tab:entity_classifier_comparision}, a special case is observed in the Inter-5 way-1 shot (Inter-51), where the MAML-ProtoNet entity classifier outperforms the SMCS entity classifier owing to two factors. (a) FewNERD's fine-grained annotations aid metric-based methods in addressing SDP, which is also elaborated in the \textbf{Ablation study} below. (b) Inter setting, distinct from Intra setting, encompasses a broader range of potential target entity types, resulting in the dispersion of type referents across the vector space, which further aids in performance. However, our SMCS entity classifier achieved comparable results, even in such situations.

(\textbf{3}) Our SMCS entity classifier surpasses MAML-ProtoNet on various datasets in both in-domain and cross-domain contexts. This advantage stems from utilizing machine common sense for referents, which effectively mitigates the SDP.

\subsection{Ablation study of entity type referents} 
\begin{table}[t]
\resizebox{\columnwidth}{!}{%
\begin{tabular}{lllllll}
\hline
\multicolumn{1}{c}{\multirow{2}{*}{Models}} & \multicolumn{2}{c}{Ontonotes} & \multicolumn{2}{c}{WNUT} & \multicolumn{2}{c}{Intra-5way} \\ \cline{2-7} 
\multicolumn{1}{c}{} & 1 shot     & 5 shot     & 1 shot     & 5 shot     & 1 shot     & 5 shot     \\ \hline
ProtoNet                & 54.70±2.1  & 77.92±0.22 & 36.21±2.02 & 54.87±2.65 & 64.91±1.38 & 79.13±0.47 \\
MAML-ProtoNet               & 67.10±0.89 & 78.05±0.45 & 50.06±1.22 & 61.88±1.13 & 67.41±0.57 & 80.30±0.35 \\
Random referents     & 56.49±1.18 & 76.08±1.21 & 49.15±1.90 & 62.01±3.00 & 59.01±1.13 & 78.11±0.93        \\
Name referents       & 53.96±1.03 & 69.63±0.58 & 50.11±0.67 & 60.66±0.83 & 59.52±0.77     & 76.34±0.71        \\
SMCS (ours)          & \textbf{69.04±0.50} & \textbf{80.17±0.64} & \textbf{62.60±1.21} & \textbf{68.68±0.91} & \textbf{69.52±0.41}        & \textbf{80.34±0.26} \\
Example referents    & 69.37±0.42 & 80.32±0.79 & 63.32±0.86 & 68.98±1.86 & 69.66±0.73        & 80.41±0.31 \\ \hline
\end{tabular}%
}
\caption{Ablation study on diverse entity type referents. We utilize ground truth entity spans without employing any span detector. Entity classifiers performances are analyzed via F1 scores. Utilizing three representative datasets: Ontonotes and FewNERD offer fine-grained entity types (18 and 66 types respectively), while WNUT provides coarse-grained entity types (6 types). Averages are based on 5 runs.}
\label{tab:ablation-study}
\end{table}
To further analyze the effectiveness of entity-type referents with machine common sense, the SMCS entity classifier was compared with entity classifiers with entity-type referents as follows: (1) Random referents, using vectors of random numbers as type referents. (2) Name referents, using embeddings of entity type names. (3) Example referents, generating entity type examples, and definitions using GPT-3.5. We also compared the SMCS entity classifier with metric-based methods, prototypical network (ProtoNet) and a meta-learning version (MAML-ProtoNet). 

In our ablation study experiments, we utilized \textbf{ground-truth entity spans} without employing any span detector. As such, the sole differentiating factor lies in the entity-type referents, which directly impact the entity classifier models. Table~\ref{tab:ablation-study} shows the ablation study results, which were analyzed from three aspects: (\textbf{A}) Models. (\textbf{B}) Number of shots. (\textbf{C}) Different datasets.

(\textbf{A}): We found that type referents generated using machine common sense (MCS) work better than other variants or classification methods. Because MCS contains proper semantic definitions of entity types and contributes to setting proper type referents. This again verifies our conclusion that MCS referents mitigate the sample dependency problem and are effective. (\textbf{Note}: obeying few-shot rules, we do not want to add extra examples, so we do not use the Example referents in the experiments in Table~\ref{tab:cross-domain-p} and Table~\ref{tab:fewnerdperformance}). The results in Table~\ref{tab:ablation-study} show that using MCS to define entity types leads to a better performance than using type names alone. In fact, our MCS referents perform similarly to example referents containing many ground-truth entity samples besides entity type definitions. The samples provided more features for classification referents and improve performance. This demonstrates that large pre-trained language models are a valuable source of classification referents.

(\textbf{B}): Based on the results from 1-shot and 5-shot cases across datasets, our SMCS entity classifier outperforms MAML-ProtoNet entity classifier, showing greater proportional improvements in 1-shot scenarios compared to 5-shot scenarios. The 1-shot case faces stronger sample dependency owing to metric-based methods, such as Prototypical Network, often utilizing the available samples to build type referents. While 5-shot cases also encounter this issue, albeit to a lesser extent. Unlike traditional methods, SMCS type referents are sample-independent, as confirmed by our results, demonstrating their efficacy.

(\textbf{C}): In this experiment, we utilized three datasets: WNUT, Ontonotes, and FewNERD, comprising 6, 18, and 66 entity types, respectively. Notably, SMCS demonstrates a greater improvement than MAML-ProtoNet on the WNUT dataset and a proportionally smaller gain on the FewNERD and Ontonotes~\cite{33} datasets. While fine-grained type annotations can potentially align few-shot samples with accurate type referents, mitigating the sample dependency problem, even in such cases, our SMCS entity classifier exhibits an effective and superior performance.

\subsection{error analysis} 
\begin{table}[t]
\centering
\resizebox{0.76\columnwidth}{!}{%
\begin{tabular}{lll}
\hline
\multirow{2}{*}{Entity types} & \multicolumn{2}{l}{F1 score} \\ \cline{2-3} 
                              & 1-shot        & 5-shot       \\ \hline
Corporation                   & 61.29±0.87    & 68.82±0.61   \\
Creative-work                 & 56.27±0.73    & 65.03±0.89   \\
Group                         & 49.42±1.68    & 61.97±0.77   \\
Location                      & 66.81±1.05    & 72.66±0.28   \\
Person                        & 69.58±2.12    & 71.39±1.31   \\
Product                       & 65.81±1.59    & 69.67±0.75   \\
All types                     & 62.60±1.21    & 68.68±0.91   \\ \hline
\end{tabular}%
}
\caption{The error analysis, focusing on entity class centroids set by machine common sense, uses ground truth entity spans to eliminate errors from false span predictions. This analysis is performed on the WNUT dataset, which pertains to the social media domain, with averages calculated over 5 runs.}
\label{tab:analysis}
\end{table}
To evaluate the impact of machine common sense on recognizing different entity types, we utilized the WNUT dataset for error analysis. This dataset, which focuses on the social media domain, is closely aligned with real-world applications and encompasses six entity types: Corporation, Creative-work, Group, Location, Person, and Product.

Analysis in Table~\ref{tab:analysis} reveals that for both 1-shot and 5-shot scenarios, performance on entities like Group and Creative-work is lower compared to others, suggesting that current Large Language Models (LLMs) are more adept at defining concrete rather than abstract concepts. Additionally, increasing support samples from 1 to 5 narrows the performance gap for Group and Creative-work, highlighting the pivotal role of support samples in improving class centroid accuracy and underscoring their significance in few-shot learning.

\section{Conclusion}
In our research, we identified the repetitive training of basic span features and the sample dependency problem within few-shot NER tasks. To address these challenges, we propose an SMCS model, structured as a pipeline. This model incorporates both a span detector and entity-type classifier. By leveraging the initialization from a Steppingstone Span Detector, SMCS effectively diminishes the repetitive training for basic span features. Moreover, by harnessing commonsense knowledge from a large language model to construct type referents, SMCS successfully mitigates the sample dependency problem, resulting in enhanced performance. Our study is the \textbf{first} instance of applying machine common sense to the realm of few-shot NER tasks. Through extensive experiments across various benchmarks, our SMCS model demonstrated superior or comparable performance to that of other strong baselines. Particularly in fine-grained few-shot NER scenarios, SMCS outperformed the strong baseline ChatGPT.

\section{Ethical Considerations}
Our contribution in this study is purely methodological, introducing a novel pipeline aimed at enhancing the performance of few-shot Named Entity Recognition (NER) and mitigating repetitive training. Consequently, there were no direct negative social impacts resulting from this contribution.

\appendices
\section{\break Dataset Introduction}
\begin{figure}[ht!]
\includegraphics[width=0.47\textwidth]{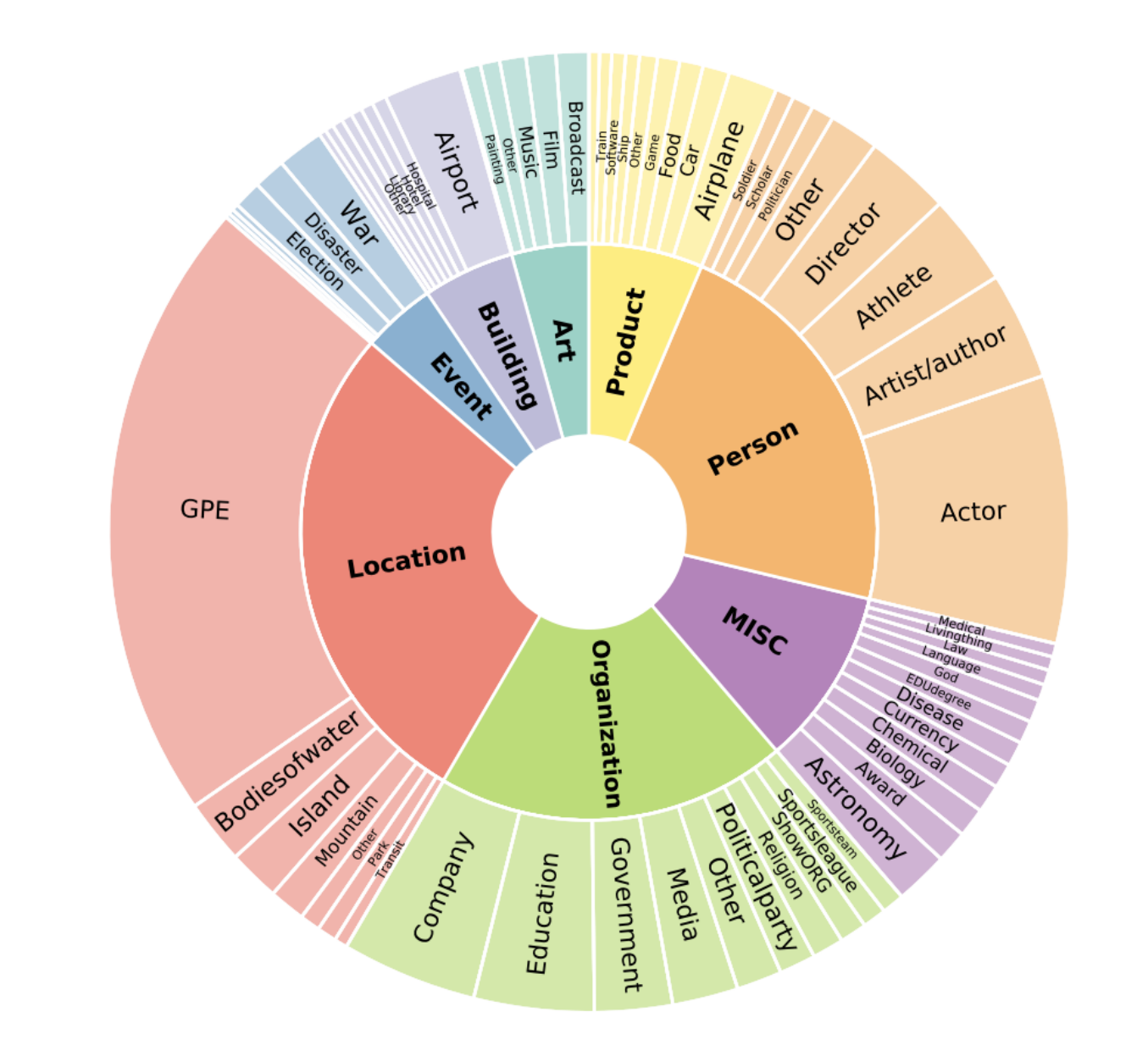}
\centering
\caption{An overview of FEW-NERD~\cite{19}. The inner circle represents the coarse-grained entity types and the
the outer circle represents the fine-grained entity types,
some types are denoted by abbreviations.}
\label{img:fewnerd_figure}
\end{figure}

Figure~\ref{img:fewnerd_figure} presents the annotation hierarchy of the Few-NERD dataset for clarity.

\section{\break Baselines Details}
\label{sec:appendix-baselines}
CONTAINER~\cite{09} used token-level contrastive learning to train the token representations of the BERT~\cite{22} model, then finetunes on the support set and applied the nearest neighbor method for evaluation.

The ESD~\cite{11} is a span-level metric-based model. It uses inter- and cross-span attention to improve the span representations.

COPNER~\cite{20} uses a novel handcrafted prompt composed of class-specific words, which serves as a supervision signal and referent in the inference stage.

MAML-ProtoNet~\cite{04} is a pipeline model that uses the meta-learning algorithm, MAML, to find a good set of parameters to adapt to new entity types.

L-TapNet+CDT~\cite{16} enhances TapNet with label semantics, pair-wise embedding, and a CDT transition mechanism.

EP-Net~\cite{10} manually sets entity type referents and disperses them to facilitate the entity classification.

Matching Networks~\cite{34} compared the similarity between query instances, support instances, and assigned entity classes.

TFP~\cite{46}, utilizes external knowledge to establish semantic anchors for each entity type, appending these anchors to input sentence embeddings as template-free prompts for in-context optimization.

2INER~\cite{43}, employs instruction finetuning to comprehend and process task-specific instructions, including both main and auxiliary tasks.

TLDT~\cite{38} modifies optimization-based meta-learning by focusing on label dependency initialization and update rules.

\section{\break Implementation Details}
We implemented our model with PyTorch 1.13.1 and trained the model using a 3090-24G GPU. For each encoder module, we used a BERT-base-uncased~\cite{22} from Huggingface as the backbone. We chose AdamW~\cite{35} as the optimizer
with a warm-up rate of 0.1. We selected five random seeds from \{171, 354, 550, 667, 985\} and reported the average results with standard deviations. The micro F1 score was used as the overall evaluation metric. The maximum sequence length was set as 128. The learning rate was $3e-5$. We used a batch size of 16 and a dropout probability of 0.1. We trained all models for 1000 steps on the Few-NERD dataset and 500 steps on the cross-domain Cross-Dataset. We set the max-loss term coefficient $\zeta$ to 2 in the training phase of the query set and 5 in other phases. We performed a \textbf{grid search} for hyper-parameters using the validation dataset. The search spaces are listed in Table~\ref{tab:search_space}:
\\
\\
\begin{table}[h!]
\begin{tabular}{lc}
\hline Learning rate & $\{1 \mathrm{e}-5,3 \mathrm{e}-5,1 \mathrm{e}-4\}$ \\
Max-loss coefficient $\zeta$ & $\{0,1,2,5,10\}$ \\
Mini-batch size & $\{8,16,32\}$ \\
\hline
\end{tabular}
\caption{The parameters search space.}
\label{tab:search_space}
\end{table}

\section{\break Acronyms Table}
We summarize all acronyms in this paper into the Table~\ref{tab:acronyms} to help reading this paper.
\begin{table*}[t]
\centering
\resizebox{\textwidth}{!}{%
\begin{tabular}{llll}
\hline
Acronyms      & Meanings                              & Acronyms         & Meanings                    \\ \hline
NER           & named entity recognition              & ESD              & an NER method               \\
LLM           & large language model                  & EP-Net           & an NER method               \\
SDP           & sample dependency problem             & MAML-ProtoNet    & an NER method               \\
SMCS          & our method in this work               & COPNER           & an NER method               \\
TLDT          & an NER method                         & TFP              & an NER method               \\
PRM           & an NER method                         & 2INER            & an NER method               \\
MANNER        & an NER method                         & ChatGPT          & a LLM from OpenAI           \\
TadNER        & an NER method                         & L-Tapnet+CDT     & an NER method               \\
MAML          & a kind of meta-learning method        & Matching Network & an NER method               \\
BIOES         & the NER span tags, B, I, O, E, S      & CONTAINER        & an NER method               \\
GPT-3.5       & a kind of LLM from OpenAI             & SSD              & Steppingstone Span Detector \\
BERT          & a language model                      & ProtoNet         & the prototypical network    \\
Few-NERD      & an in-domain NER datasets collection   & WNUT             & an NER dataset              \\
Cross-Dataset & a cross-domain NER datasets collection & GUM              & an NER dataset              \\
CoNLL-03      & an NER dataset                        & OntoNotes        & an NER dataset              \\ \hline
\end{tabular}%
}
\caption{Summary of Acronyms. This table provides a comprehensive list of all acronyms used in the document, along with their corresponding meanings.}
\label{tab:acronyms}
\end{table*}

\section{\break Prompts for ChatGPT}
Here is a sample prompt tailored for ChatGPT, specifically intended for the CoNLL NER dataset~\cite{30} 1-shot case. This prompt format can be adjusted to accommodate other NER datasets such as OntoNotes, FewNERD, and similar ones. Notably, the key variation between datasets lies in their unique \textbf{entity lists}.

The prompt consists of three sections: definition, few-shot samples, and query requests.

\textbf{Definition}: We have the following entity types in the \textbf{entity type list} ['LOC', 'MISC', 'ORG', 'PER']. We want to annotate each word in the sentence using the above entity types. If a word does not belong to the above entity types, we label it using the entity tag 'O'. We will provide some sentences and their corresponding entity type label sequences as examples to improve your understanding.

\textbf{Few-shot Samples}: The examples are as follows,[(['viktoria', 'zizkov', '3', '0', '1', '2', '3', '8', '1'], ['ORG', 'ORG', 'O', 'O', 'O', 'O', 'O', 'O', 'O']), (['tennis', 'monday', 's', 'results', 'from', 'us', 'open'], ['O', 'O', 'O', 'O', 'O', 'MISC', 'MISC']), (['11776', '12442'], ['O', 'O']), (['tanya', 'dubnicoff', 'canada', 'beat', 'michelle', 'ferris', 'australia', '20'], ['PER', 'PER', 'LOC', 'O', 'PER', 'PER', 'LOC', 'O']), (['mountain', 'view', 'calif', '19960825'], ['LOC', 'LOC', 'LOC', 'O'])].

\textbf{Query Request}: Now we have the following query sentences, please label sentences with entity types ['LOC', 'MISC', 'ORG', 'PER'] or 'O' tag.(['1', 'michael', 'schumacher', 'germany', 'ferrari', '1', 'hour', '28', 'minutes'], ['times', 'in', 'seconds'], ['drawn', 'lost', 'goals', 'for', 'against', 'points'], ['split', 'croatia', '9', '4', '5', '13'], ['71', '64'], ['basketball', 'tournament', 'on', 'friday'], ['the', 'plan', 'has', 'raised', 'a', 'storm', 'of', 'protest', 'from', 'us', 'avocado', 'growers', 'who', 'are', 'largely', 'concentrated', 'in', 'california'], ['boston', '19960829'], ['12152012', '1200m', '525', '530'], ['jordan', 'said', 'a', 'small', 'group', 'of', 'developing', 'nations', 'that', 'oppose', 'linking', 'trade', 'talks', 'and', 'labour', 'conditions', 'had', 'pressured', 'world', 'trade', 'organisation', 'wto', 'officials', 'to', 'prevent', 'hansenne', 'from', 'taking', 'the', 'platform', 'to', 'urge', 'such', 'links'], ['mike', 'cito', '17', 'was', 'expelled', 'from', 'st', 'pius', 'x', 'high', 'school', 'in', 'albuquerque', 'after', 'an', 'october', 'game', 'in', 'which', 'he', 'used', 'the', 'sharpened', 'chin', 'strap', 'buckles', 'to', 'injure', 'two', 'opposing', 'players', 'and', 'the', 'referee'], ['8', 'martin', 'keino', 'kenya', '73888'], ['more', 'than', '10', 'weapons', 'including', 'automatic', 'kalashnikov', 'rifles', 'were', 'stolen', 'from', 'an', 'arms', 'store', 'in', 'belgium', 'police', 'said', 'on', 'saturday'], ['cricket', 'pakistan', '3394', 'v', 'england', 'close'], ['anton', 'ferreira'], ['we', 'hope', 'for', 'the', 'best', 'that', 'it', 'really', 'has', 'ended', 'so', 'we', 'can', 'live', 'in', 'peace'], ['15', 'retief', 'goosen', 'south', 'africa', '188143'], ['seoul', '19960823'], ['analysts', 'hold', 'dutch', 'ptt', 'estimates'], ['stocks', 'to', 'watch']). Each sentence result should have the following format: ([sentence words], [sentence words entity labels]). For each sentence the prediction is a Python tuple,the first element is a Python list containing sentence words, and the second element is the corresponding entity label for each sentence word. The results should be several tuples separated by a single comma character.

\section*{Acknowledgment}
We express our heartfelt gratitude to all those who took the time to read this paper and offer insightful comments. Additionally, we wish to express our appreciation to our families and friends whose unwavering support has been invaluable throughout this endeavor.

\bibliography{citation_resource}
\bibliographystyle{IEEEtran}

\begin{IEEEbiography}[{\includegraphics[width=1in,height=1.25in,clip,keepaspectratio]{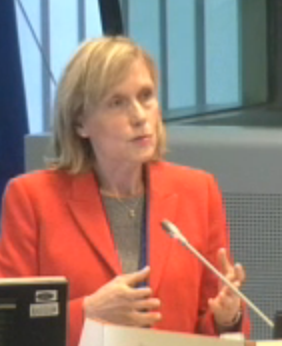}}]{Marie-Francine (Sien) Moens} holds a MSc in Computer Science and a PhD degree in Computer Science from KU Leuven. She is a Full Professor at the KU Leuven. In 2021 she was the general chair of the 2021 Conference on Empirical Methods in Natural Language Processing (EMNLP 2021). In 2011 and 2012 she was appointed as chair of the European Chapter of the Association for Computational Linguistics (EACL) and was a member of the executive board of the Association for Computational Linguistics (ACL). She is currently associate editor of the journal IEEE Transactions on Pattern Analysis and Machine Intelligence (TPAMI) and was a member of the editorial board of the journal Foundations and Trends in Information Retrieval from 2014 till 2018.
She is holder of the ERC Advanced Grant CALCULUS (2018-2024) granted by the European Research Council. From 2014 till 2018 she was the scientific manager of the EU COST action iV\&L Net (The European Network on Integrating Vision and Language). She is a fellow of the European Laboratory for Learning and Intelligent Systems (ELLIS). Her research topics include: machine learning for natural language processing and the joint processing of language and visual data, deep learning, Multimodal and multilingual processing, machine learning models for structured prediction and generation, and so on.
\end{IEEEbiography}

\begin{IEEEbiography}[{\includegraphics[width=1in,height=1.25in,clip,keepaspectratio]{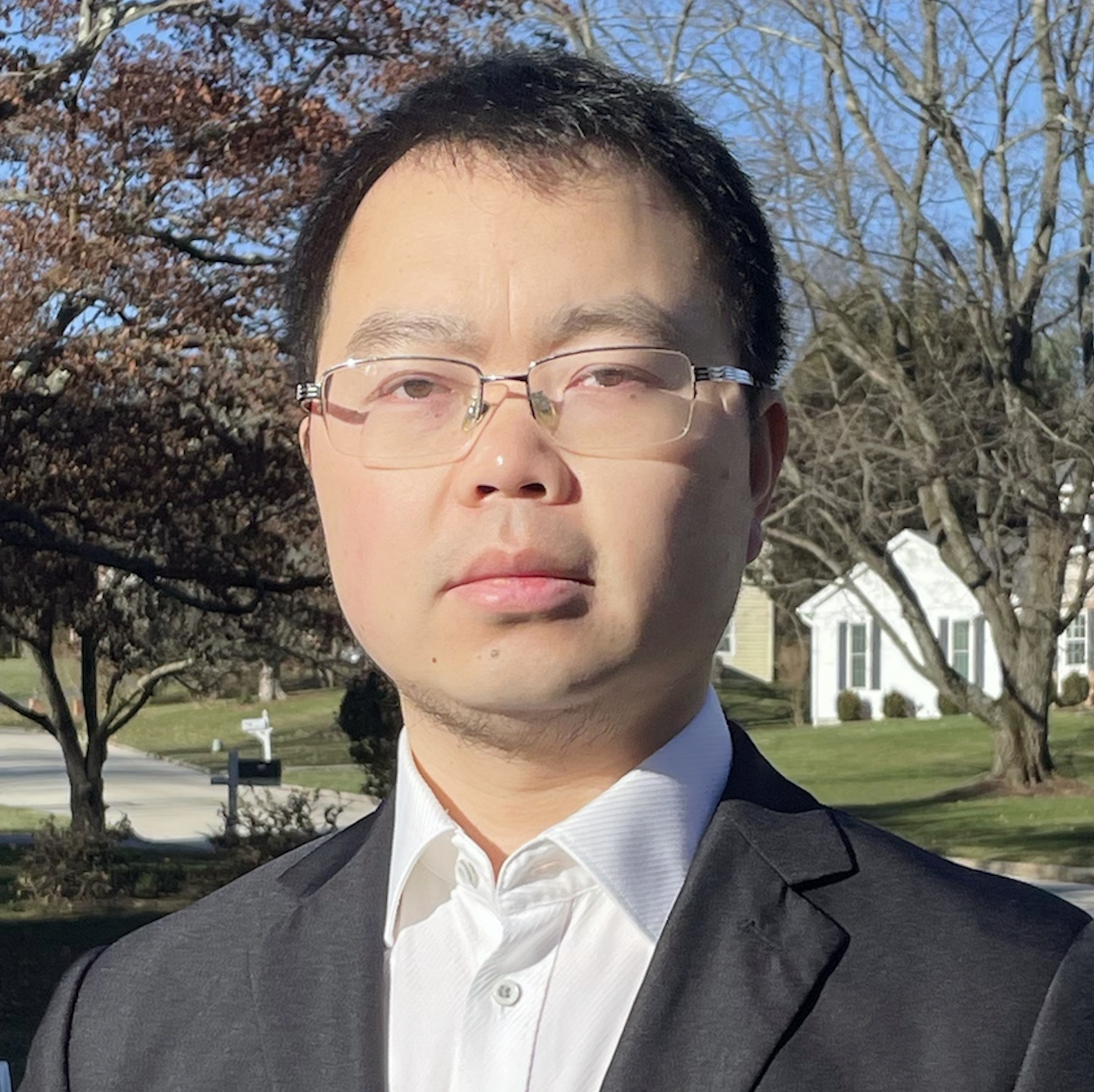}}]{Wenpeng Yin} holds an MSc in Computer Science from Peking University, and a Ph.D. degree in Computer Science from Ludwig-Maximilians-Universität München (LMU Munich). He is an Assistant Professor in the Department of Computer Science and Engineering at the Pennsylvania State University (PSU), leading the AI4Research lab. Prior to joining PSU, Dr. Yin worked as a postdoctoral researcher at the University of Pennsylvania and as a senior research scientist at Salesforce Research. He was the co-chair of Wise-Supervision in 2022, and Senior Area Chair of NAACL'2021, ACL Rolling Review (ARR), IJCNLP-AACL'23, LREC-COLING'24, 
 and EACL'24. His research interests span Natural Language Processing, Multimodality, Human-Centered AI, etc.
\end{IEEEbiography}


\begin{IEEEbiography}[{\includegraphics[width=1in,height=1.25in,clip,keepaspectratio]{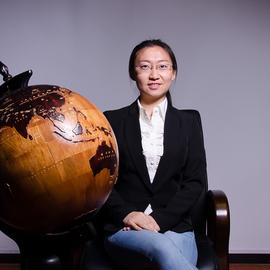}}]{Dan Li} holds an MSc in Computer Science from Dalian University of Technology, and a PhD degree in Computer Science from University of Amsterdam. She is a data scientist at Elsevier working on machine learning and deep learning. She is a member of the European Laboratory for Learning and Intelligent Systems (ELLIS). Her research interests include: conversational search, crowdsourcing label denoising, information retrieval, probabilistic graphical models, gaussian processes.
\end{IEEEbiography}

\begin{IEEEbiography}[{\includegraphics[width=1in,height=1.25in,clip,keepaspectratio]{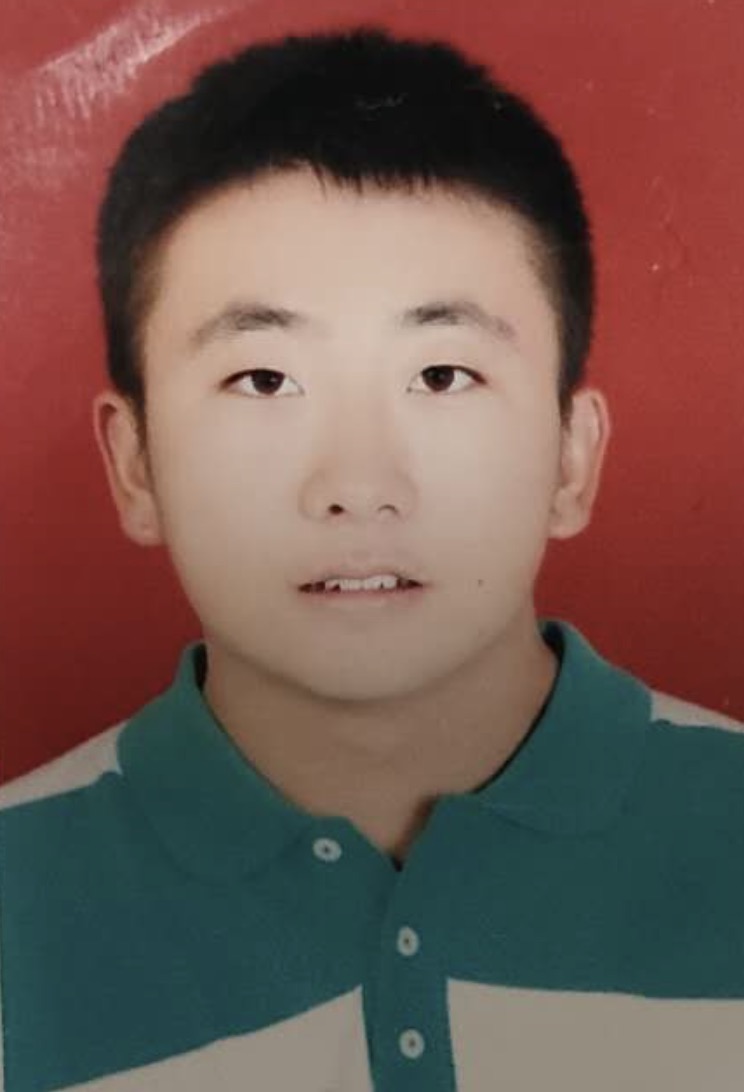}}]{Chang Tian} holds a B.Eng in Electronic Engineering from the Shandong University and an MSc in Artificial Intelligence from the University of Amsterdam. He is a PhD candidate in Computer Science from KU Leuven. He serves as a reviewer for conferences or journals SIGKDD, ACL, TPAMI, and CCL. His research interests include: deep learning, machine learning, natural language processing, dialogue systems, few-shot learning, and information extraction.
\end{IEEEbiography}

\EOD

\end{document}